\documentclass{article}
\usepackage[margin=1in]{geometry}

\usepackage{amssymb}
\usepackage{amsmath}
\usepackage{amsthm}
\usepackage{graphicx}
\usepackage{xcolor}

\usepackage[inline]{enumitem}
\usepackage{booktabs}
\usepackage{multicol}

\begin{document}

\title{Learning Generalizable Neural Operators for Inverse Problems}
\author{Adam J. Thorpe, Stepan Tretiakov, Dibakar Roy Sarkar, Krishna Kumar, Ufuk Topcu}
\maketitle

\begin{abstract}
Inverse problems challenge existing neural operator architectures because ill-posed inverse maps violate continuity, uniqueness, and stability assumptions. We introduce B2B${}^{-1}$, an inverse basis-to-basis neural operator framework that addresses this limitation. Our key innovation is to decouple function representation from the inverse map. We learn neural basis functions for the input and output spaces, then train inverse models that operate on the resulting coefficient space. This structure allows us to learn deterministic, invertible, and probabilistic models within a single framework, and to choose models based on the degree of ill-posedness. We evaluate our approach on six inverse PDE benchmarks, including two novel datasets, and compare against existing invertible neural operator baselines. We learn probabilistic models that capture uncertainty and input variability, and remain robust to measurement noise due to implicit denoising in the coefficient calculation. Our results show consistent re-simulation performance across varying levels of ill-posedness. By separating representation from inversion, our framework enables scalable surrogate models for inverse problems that generalize across instances, domains, and degrees of ill-posedness.
\end{abstract}

\section{Main}

Inverse problems seek to uncover hidden causes from observable effects, but are often ill-posed: solutions may be non-unique and small perturbations in the data can lead to large, unstable variations in the output \cite{Kabanikhin+2008+317+357, doi:10.1137/1021044}.
Despite this difficulty, inverse problems are fundamental to scientific inquiry, from seismic imaging in geophysics \cite{o2023imaging} to noninvasive medical diagnosis \cite{song2022solving}.
In practice, these problems are solved by iterative algorithms or hand-crafted regularization techniques, which are computationally expensive and require domain experts to tune parameters for every new observation.
Machine learning offers an attractive path to construct fast, scalable surrogate models that learn underlying patterns from data and provide near-instantaneous predictions.
However, the ill-posedness of inverse problems, such as non-uniqueness and ill-conditioning, directly conflicts with the assumptions underlying most machine learning methods, which typically presume a continuous, well-defined mapping from inputs to outputs \cite{doi:10.1137/23M1586872, doi:10.1137/23M1568739, koch2021instability, doi:10.1073/pnas.1907377117}.
This highlights a central open challenge for inverse problems: to learn fast, general-purpose surrogate models from data. 

Current computational tools fail to provide generalizable surrogates.
Existing approaches to learning inverse solutions are specialized rather than general, meaning they solve individual instances but do not provide scalable surrogates that generalize across problem settings. 
Physics-informed neural networks (PINNs) combine data with governing equations \cite{RAISSI2019686}, but they act as instance-specific solvers rather than general operators; because a PINN must be retrained for every new problem configuration or boundary condition, it cannot function as a fast surrogate across a domain \cite{torres2025adaptive}.
Reduced-order modeling (ROM) achieves tractability by projecting dynamics onto low-dimensional manifolds \cite{doi:10.1137/12089586X, 6468442, Ghattas_Willcox_2021, doi:10.1137/090775622, Stuart_2010}, but the choice of basis is usually dependent on a specific system or geometry, which limits generalization. 

Operator learning, by contrast, seeks to learn transformations between entire function spaces instead of a single input-output pair, offering a path toward generalizable surrogate models. 
Forward-trained operator learning architectures, such as DeepONet \cite{lu2021learning}, Fourier Neural Operators (FNO) \cite{li2021fourier}, and their variants \cite{10.1145/3648506, goswami2023physics, kontolati2024learning, BAHMANI2025118113, tretiakov2025setonet}, have achieved considerable success as surrogate models for forward problems, but inverse maps violate the continuity and uniqueness assumptions these models were designed to exploit \cite{doi:10.1073/pnas.1907377117, doi:10.1137/23M1568739, doi:10.1137/23M1586872}. 
Recent DeepONet and FNO variants enable strict invertibility \cite{kaltenbach2023invertibleNeuralOperators, pmlr-v258-long25a, pmlr-v202-molinaro23a}, but rely on restrictive architectural constraints such as matched input/output dimensions and sensitive training schemes that limit flexibility.
While these models allow inverse evaluation, enforcing a one-to-one structure on inherently one-to-many problems forces the model to ignore ambiguity and uncertainty, leading to numerical instability or the collapse of meaningful variability.

We introduce B2B${}^{-1}$, a modular framework for learning inverse neural operators that reframes inversion as distributional inference over function spaces. Unlike classical neural operators that regress a single output from a given input, inverse problems require reasoning over entire sets of plausible inputs conditioned on noisy, indirect observations. To address this, we decouple representation learning from inversion by extending basis-to-basis (B2B) operator learning \cite{INGEBRAND2025117646} to the inverse setting. First, we learn neural basis representations of the input and output function spaces. Then, we train an inverse map between coefficient spaces that specifically tackles the challenges associated with ill-posed problems. 
This separation enables us to: 
\begin{enumerate*}[label=(\arabic*)]
    \item 
    model uncertainty and non-uniqueness in ill-posed problems via probabilistic inverse maps;
    \item 
    recover bijective relations when the relation is one-to-one with invertible networks; and
    \item
    solve well-posed problems efficiently using deterministic models.
\end{enumerate*}
We evaluate these inverse models on a diverse set of benchmark PDE operator learning problems, 
and compare our approach against existing invertible Fourier neural operator learning approaches \cite{pmlr-v258-long25a}.

\begin{figure}[!t]
    \centering
    \includegraphics[width=\linewidth]{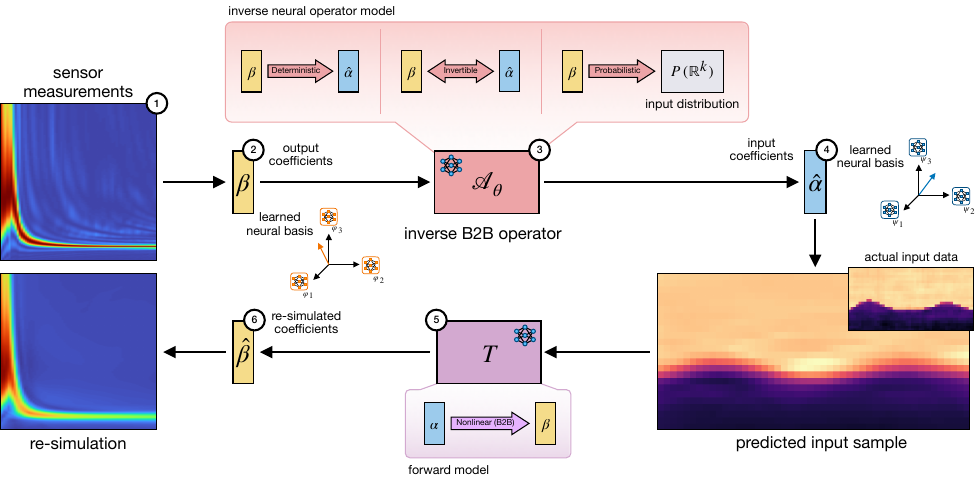}
    \caption{B2B${}^{-1}$ inverse operator framework. 
    Sensor measurements (1) are projected onto learned neural basis functions to produce a low-dimensional coefficient vector $\beta$ (2). The inverse neural operator maps $\beta$ to input coefficients $\alpha$ (2-4) of a second learned neural basis of the input space. This operator can be deterministic, invertible, or probabilistic; in the probabilistic case, it defines a distribution over plausible $\alpha$. We pass sampled inputs through a separate forward neural operator $T$ for re-simulation (5–6) and validation.
    }
    \label{fig: inverse b2b diagram}
\end{figure}

\section{Results}

\subsection{Reframing Inverse Neural Operator Learning}

Classical inverse problems seek to recover a function $f$ given noisy observations $h(y_i) = T(f(x_i)) + \epsilon_i$, where $T : \mathcal{G} \to \mathcal{H}$ is a (possibly nonlinear) operator between spaces $\mathcal{G}$ and $\mathcal{H}$, and $\epsilon_i$ denotes measurement noise.
Even when $T$ is known and deterministic, the inverse may be non-unique, unstable, or undefined \cite{Kabanikhin+2008+317+357}. 
Standard neural operators such as DeepONet \cite{lu2021learning}, Fourier Neural Operators (FNO) \cite{li2021fourier}, and current B2B formulations \cite{INGEBRAND2025117646}, are trained on forward mappings from $f$ to $h$.
These models produce deterministic outputs and do not explicitly address the structural challenges of inverse learning.

To address this, we define an inverse neural operator as a map from observations to probability distributions over the input functions, $\mathcal{A}_{\theta} : \mathcal{H} \to \mathcal{P}(\mathcal{G})$, where $\mathcal{P}(\mathcal{G})$ denotes distributions over the input space. 
This formulation introduces two changes to prior work in neural operators:
\begin{enumerate*}[label=(\arabic*)]
     \item 
     we treat the inverse operator as probabilistic, capturing non-uniqueness and uncertainty, and
     \item 
     we recast inversion as a function-space learning problem.
\end{enumerate*}
Given observations from an output function $h$, the goal is to infer a distribution over input functions $p \in \mathcal{P}(\mathcal{G})$. The goal is to produce plausible input reconstructions whose forward evaluations are consistent with the observations.
Deterministic models are a special case, where the prediction is a point estimate $f \in \mathcal{G}$ that typically corresponds to the conditional mean under a squared-error loss. 

To implement this probabilistic formulation, we extend the B2B operator learning architecture \cite{INGEBRAND2025117646} to the inverse setting. 
The key idea is to decouple function representation from inversion by learning finite, data-driven basis representations for both the input and output function spaces.
This structure allows us to instantiate inverse maps tailored to the ill-posedness of the problem. 
The method separates function representation from operator learning and learns neural basis functions independently for inputs and outputs. After projecting functions into their respective coefficient spaces, the inverse learning task becomes mapping output coefficients $\beta \in \mathbb{R}^{l}$ to distributions over input coefficients $\alpha \in \mathbb{R}^{k}$. 

To instantiate the basis-to-basis inverse neural operator learning framework, B2B${}^{-1}$, we implement multiple inverse models, each reflecting different assumptions about the inverse map: 
\begin{itemize}
    \item 
    Deterministic models, including linear regressors and MLPs, assume a unique, stable mapping from outputs to inputs.
    \item 
    Strictly invertible models, such as INNs and cINNs, impose a bijective structure and learn forward and inverse maps jointly. 
    \item 
    Probabilistic models map outputs to full input distributions. These include conditional variational autoencoders (cVAE), which use latent variables to encode variability; mixture density networks (MDN), which represent multimodal outputs as Gaussian mixtures; and conditional normalizing flows (RealNVP), which learn expressive, invertible transformations over probability densities.
\end{itemize}
All models operate on the same learned coefficient representations, isolating the effect of model class. We also compare against iFNO, a fully invertible neural operator trained directly on discretized function data~\cite{pmlr-v258-long25a}.

\subsection{Performance and Comparison}

We evaluate inverse neural operators on six PDE-based inverse problems selected to span a range of inverse structures, dimensionalities, and degrees of ill-posedness.
These include three standard synthetic benchmarks: Darcy flow, Burgers' equation, and elastic plate deformation. 
Darcy flow provides a nearly bijective inverse under fixed boundary conditions and serves as a well-posed baseline. Elastic plate deformation introduces moderate, localized ambiguity due to stress concentrations from irregular geometry. Burgers' equation is strongly ill-posed, as diffusion and shock formation cause many distinct initial conditions to evolve into nearly identical outputs.
We also introduce two new benchmarks designed to test inverse operator models under function-space ambiguity and high-frequency structure: Chladni plate resonance and wave scattering. 
The Chladni dataset exhibits strong multimodality, where distinct forcing patterns produce nearly identical resonance fields. 
The wave scattering dataset requires recovering fine-scale geometric features from far-field observations. 
Finally, the full waveform inversion (FWI) dataset represents a canonical geophysics inverse problem. Traditional FWI approaches require extensive problem-specific tuning. A general-purpose inverse operator that can approximate solutions directly from data would represent a significant practical advance.
Technical details of each benchmark are provided in the Supplementary Results \ref{section: supplementary results benchmarks}.

Because inverse solutions are generally non-unique on ill-posed problems, direct comparison between predicted and ground-truth inputs is not meaningful.
Instead, we evaluate model performance using a forward-consistency metric: for each predicted input $\hat{f}$, we apply a separately trained B2B forward model $\hat{T}$ to generate a re-simulated output $\hat{h} = \hat{T}(\hat{f})$, and compute the relative $L_2$ discrepancy between $\hat{h}$ and the observed measurement $h$. This approach provides a consistent, task-agnostic evaluation criterion of validity for the predicted inputs. See \S \ref{section: resimulation error} for additional technical details.

Across these benchmarks, inverse model performance depends primarily on the structure of the inverse mapping. Deterministic models achieve low re-simulation error on well-posed and low-ambiguity problems but collapse to mean or trivial solutions as ill-posedness increases. Invertible models rely on approximate bijectivity and exhibit similar degradation when this assumption fails. Probabilistic models are necessary to represent non-uniqueness and recover diverse, forward-consistent inputs on strongly ill-posed tasks. Table \ref{tab: results} summarizes quantitative performance across models and datasets.

\begin{table*}[!t]
\scriptsize
\centering
\begin{tabular*}{\textwidth}{lcccccc}
\toprule
 & Burgers & Darcy Flow & Chladni & Elastic & Wave Scattering & FWI \\
\midrule
\multicolumn{7}{l}{\textbf{Dataset Parameters}} \smallskip \\
Input Dimension & [128] & [101] & [25, 25] & [101] & [100] & [24, 48] \\
Output Dimension & [128] & [101] & [25, 25] & [1048] & [200, 200] & [400, 76] \\
Training Size & 10,000 & 10,000 & 10,000 & 1,900 & 1,000 & 80,000 \\
\midrule
\multicolumn{7}{l}{\textbf{Function Encoder and B2B Forward Model Relative $L_2$ Error}} \smallskip \\
Input FE   
  & $7.58 \pm 0.13 \mathrm{e}^{-3}$ 
  & $9.46 \pm 0.35 \mathrm{e}^{-3}$ 
  & $1.52 \pm 0.02 \mathrm{e}^{-2}$ 
  & $2.25 \pm 0.34 \mathrm{e}^{-2}$ 
  & $2.13 \pm 0.09 \mathrm{e}^{-2}$
  & $7.22 \pm 0.04 \mathrm{e}^{-2}$ \\
Output FE  
  & $4.57 \pm 0.12 \mathrm{e}^{-3}$ 
  & $6.93 \pm 0.31 \mathrm{e}^{-3}$ 
  & $8.53 \pm 0.15 \mathrm{e}^{-3}$ 
  & $5.44 \pm 0.28 \mathrm{e}^{-3}$ 
  & $2.25 \pm 0.05 \mathrm{e}^{-1}$
  & $5.54 \pm 0.17 \mathrm{e}^{-2}$ \\
B2B (Linear)     
  & $9.67 \pm 7.37 \mathrm{e}^{-1}$ 
  & $1.94 \pm 3.75 \mathrm{e}^{1}$ 
  & $3.44 \pm 1.92 \mathrm{e}^{-2}$ 
  & $25.14 \pm 14.89$ 
  & $2.08 \pm 1.97 \mathrm{e}^{2}$
  & $2.29 \pm 0.00 \mathrm{e}^{-1}$ \\
B2B (Nonlinear)  
  & $1.65 \pm 0.99 \mathrm{e}^{-2}$ 
  & $4.57 \pm 1.74 \mathrm{e}^{-2}$ 
  & $5.09 \pm 2.18 \mathrm{e}^{-2}$ 
  & $5.24 \pm 2.99 \mathrm{e}^{-2}$ 
  & $2.67 \pm 0.16 \mathrm{e}^{-1}$
  & $2.25 \pm 0.15 \mathrm{e}^{-1}$ \\
\midrule
\multicolumn{7}{l}{\textbf{Inverse Model Re-simulation Relative $L_2$ Error}} \smallskip \\
Linear 
  & $5.83 \pm 5.64 \mathrm{e}^{-1}$ 
  & $4.15 \pm 1.25 \mathrm{e}^{-1}$ 
  & $4.91 \pm 2.07 \mathrm{e}^{-2}$
  & $75.29 \pm 79.00$
  & $3.15 \pm 0.96 \mathrm{e}^{-1}$ 
  & $1.82 \pm 0.05 \mathrm{e}^{-1}$ \\
Linear-Inv 
  & $1.58 \pm 0.67$ 
  & $5.38 \pm 5.09$ 
  & $3.76 \pm 6.83 \mathrm{e}^{3}$
  & $1.10 \pm 0.11$
  & $7.32 \pm 5.32$ 
  & $1.02 \pm 1.09 \mathrm{e}^{2}$ \\
Nonlinear (MLP) 
  & $4.73 \pm 1.32 \mathrm{e}^{-2}$ 
  & $5.58 \pm 1.89 \mathrm{e}^{-2}$ 
  & $7.99 \pm 1.41 \mathrm{e}^{-2}$
  & $5.63 \pm 2.68 \mathrm{e}^{-2}$
  & $2.66 \pm 0.16 \mathrm{e}^{-1}$ 
  & $2.07 \pm 0.05 \mathrm{e}^{-1}$ \\
\midrule
INN-Additive 
  & $1.50 \pm 0.90 \mathrm{e}^{-1}$ 
  & $1.46 \pm 0.62 \mathrm{e}^{-1}$ 
  & $1.10 \pm 0.06 \mathrm{e}^{-1}$
  & $1.27 \pm 0.66 \mathrm{e}^{-1}$
  & $3.01 \pm 0.17 \mathrm{e}^{-1}$ 
  & $3.10 \pm 0.28 \mathrm{e}^{-1}$ \\
cINN-Additive 
  & $5.63 \pm 2.00 \mathrm{e}^{-2}$ 
  & $9.99 \pm 2.88 \mathrm{e}^{-2}$ 
  & $8.32 \pm 1.29 \mathrm{e}^{-2}$
  & $6.97 \pm 1.98 \mathrm{e}^{-2}$
  & $2.66 \pm 0.17 \mathrm{e}^{-1}$ 
  & $2.60 \pm 0.20 \mathrm{e}^{-1}$ \\
INN-Affine 
  & $7.93 \pm 2.62 \mathrm{e}^{-1}$ 
  & $1.69 \pm 0.68$ 
  & $8.03 \pm 0.30 \mathrm{e}^{-1}$
  & $1.24 \pm 0.18$
  & $6.55 \pm 0.07 \mathrm{e}^{-1}$ 
  & $5.53 \pm 0.45 \mathrm{e}^{-1}$ \\
cINN-Affine 
  & $5.03 \pm 1.32 \mathrm{e}^{-2}$ 
  & $8.21 \pm 1.14 \mathrm{e}^{-2}$ 
  & $1.07 \pm 0.09 \mathrm{e}^{-1}$
  & $7.33 \pm 1.68 \mathrm{e}^{-2}$
  & $2.67 \pm 0.17 \mathrm{e}^{-1}$ 
  & $2.64 \pm 0.18 \mathrm{e}^{-1}$ \\
\midrule
cVAE 
  & $8.04 \pm 0.86 \mathrm{e}^{-2}$ 
  & $9.71 \pm 1.09 \mathrm{e}^{-2}$ 
  & $9.70 \pm 1.18 \mathrm{e}^{-2}$
  & $6.28 \pm 2.40 \mathrm{e}^{-2}$
  & $7.17 \pm 0.87 \mathrm{e}^{-2}$ 
  & $2.09 \pm 0.08 \mathrm{e}^{-1}$ \\
MDN 
  & $3.28 \pm 0.24 \mathrm{e}^{-1}$ 
  & $3.50 \pm 0.91 \mathrm{e}^{-1}$ 
  & $5.10 \pm 0.10 \mathrm{e}^{-1}$
  & $9.31 \pm 1.34 \mathrm{e}^{-2}$
  & $2.70 \pm 0.16 \mathrm{e}^{-1}$ 
  & $8.54 \pm 0.82 \mathrm{e}^{-1}$ \\
cINN-Additive-Prob 
  & $1.06 \pm 0.30 \mathrm{e}^{-1}$
  & $2.64 \pm 0.16 \mathrm{e}^{-1}$
  & $9.33 \pm 1.32 \mathrm{e}^{-2}$
  & $7.96 \pm 1.99 \mathrm{e}^{-2}$
  & $2.68 \pm 0.15 \mathrm{e}^{-1}$
  & $2.87 \pm 0.08 \mathrm{e}^{-1}$ \\
RealNVP 
  & $8.51 \pm 3.67 \mathrm{e}^{-2}$ 
  & $1.56 \pm 0.30 \mathrm{e}^{-1}$ 
  & $1.55 \pm 0.12 \mathrm{e}^{-1}$
  & $1.10 \pm 0.51 \mathrm{e}^{-1}$
  & $2.67 \pm 0.16 \mathrm{e}^{-1}$ 
  & $2.52 \pm 0.10 \mathrm{e}^{-1}$ \\
\midrule 
iFNO 
  & $7.26 \pm 0.54 \mathrm{e}^{-2}$
  & $2.70 \pm 0.37 \mathrm{e}^{-3}$
  & --
  & $1.93 \pm 0.17 \mathrm{e}^{-1}$
  & $9.22 \pm 0.05 \mathrm{e}^{-1}$
  & -- \\
\bottomrule
\end{tabular*}
\caption{Comparison of $\text{B2B}^{-1}$ and iFNO model performance across PDE inverse problems. We report re-simulation error for each inverse model across benchmark PDE datasets across 5 independent seeds. Bold values indicate the best-performing model for each task. Errors are measured as relative $L_2$ error between the forward-simulated output values and the original observation, using predicted inputs. Dataset parameters and forward model performance (relative $L_{2}$ error) are included for reference. Entries marked as ``--'' indicate cases where the metric was undefined due to numerical instability or exceeded a predefined error threshold, and are therefore omitted. }
\label{tab: results}
\end{table*}

\subsubsection{Performance on Classical PDE Benchmarks}

\begin{figure}[!t]
    \centering
    \includegraphics[width=\linewidth]{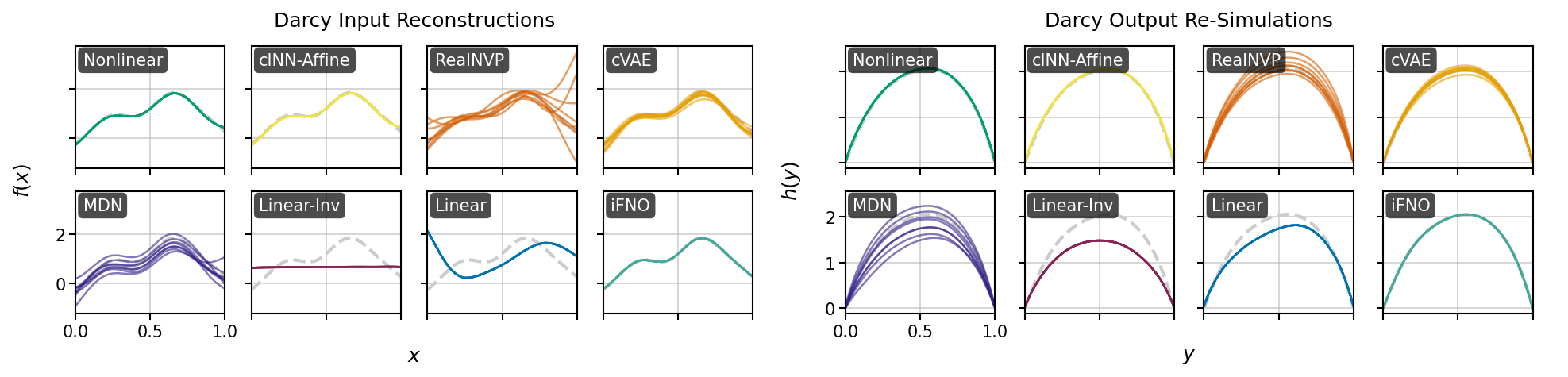}
    \includegraphics[width=\linewidth]{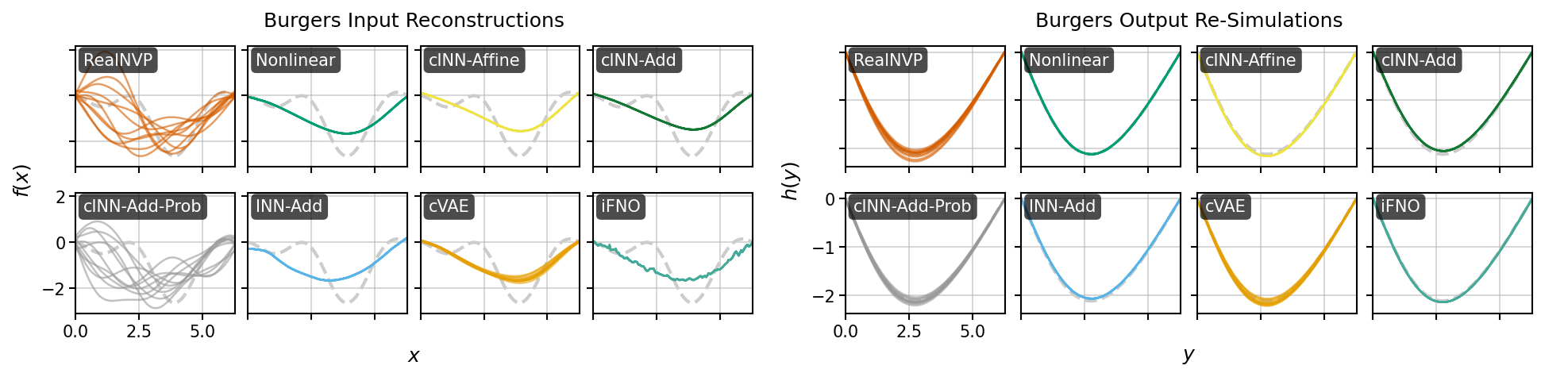}
    \includegraphics[width=\linewidth]{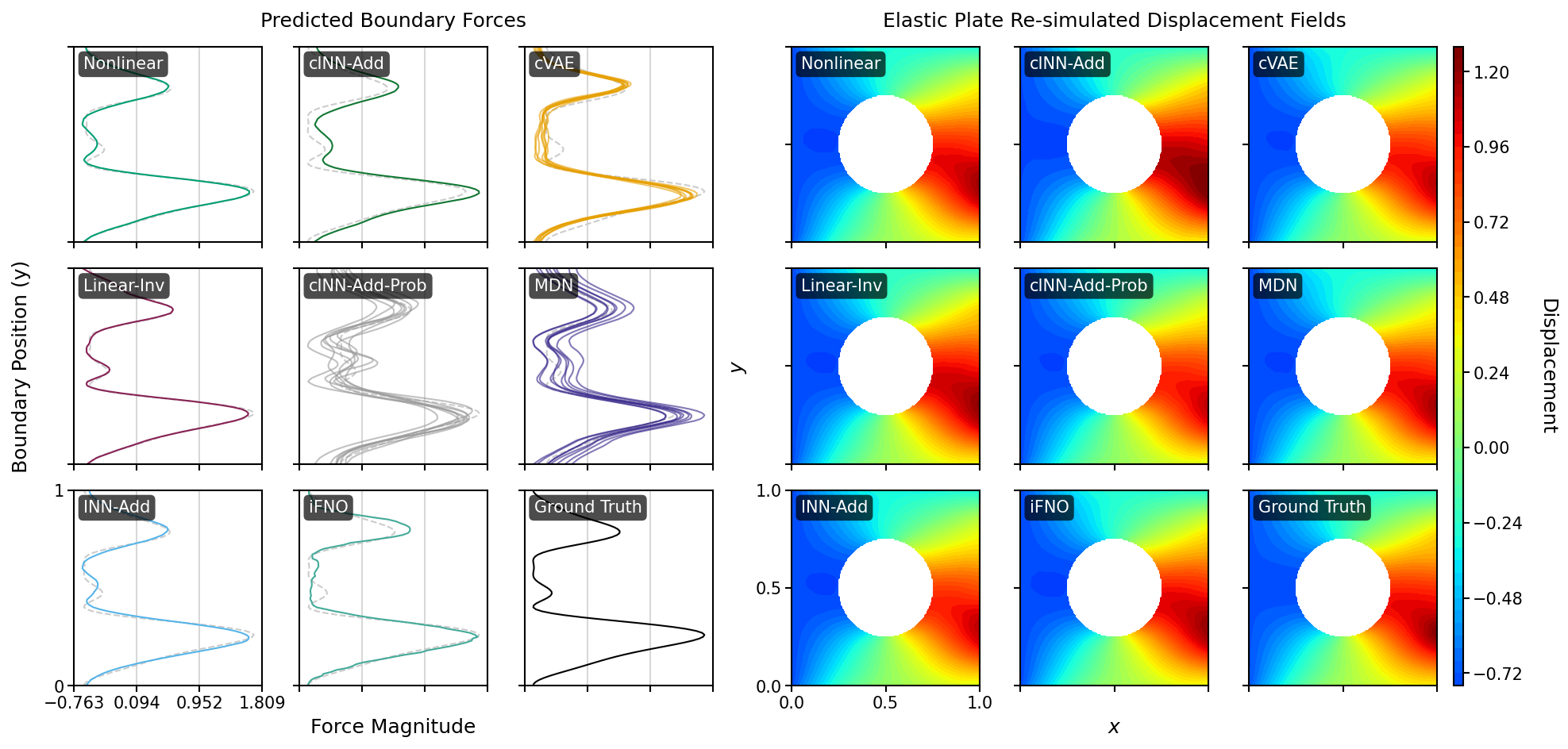}
    \caption{Qualitative comparison of inverse model performance on classical PDE benchmarks. We visualize the best average representative sample across test data for each dataset (Darcy, Burgers, Elastic Plate). For probabilistic models, input reconstructions show random samples drawn from the predicted distribution. The corresponding output re-simulations, generated by passing these samples through the forward model, are shown on the right. Effective models recover plausible input functions and produce accurate output reconstructions when evaluated under the forward operator.
    }
    \label{fig: classic pde results}
\end{figure}

We first evaluate inverse neural operators on three classical PDE benchmarks: Darcy flow, elastic plate deformation, and Burgers' equation. These benchmarks are widely used in operator learning, and provide controlled settings where the structure of the inverse map is well understood. Figure \ref{fig: classic pde results} compares input predictions and re-simulations on a representative sample from each dataset.

Darcy flow models steady-state pressure in porous media, governed by an elliptic PDE 
$-\frac{d}{dx}(\kappa(s(x))\,\frac{ds(x)}{dx}) = u(x)$,
where $\kappa(x)$ is the unknown permeability field. Under appropriate boundary conditions, the mapping from pressure to permeability is nearly one-to-one but nonlinear. 
Figure \ref{fig: classic pde results} shows that deterministic models including the nonlinear MLP, cINN, and iFNO recover accurate permeability fields and achieve low re-simulation error. This behavior is consistent with the near-bijective structure of the inverse problem. In contrast, linear models perform poorly, often producing trivial solutions, such as constant permeability fields. Probabilistic models learn accurate mean permeability fields but retain unnecessary variance, retaining uncertainty that is not intrinsic to the inverse mapping.

The elastic plate benchmark models in-plane deformation of a thin rectangular plate under uniaxial tension. 
The inverse problem is to infer the spatially varying force $w(x, y)$ from observations of the displacement field.
The presence of a central hole introduces localized stress concentrations, leading to moderate ill-posedness confined to specific spatial regions rather than global non-uniqueness.
Figure \ref{fig: classic pde results} shows that the nonlinear MLP achieves the lowest re-simulation error, while the linear inverse model also performs well. This indicates that the inverse mapping is locally linear in the learned B2B coefficient space. Probabilistic models exhibit behavior similar to Darcy flow. The cVAE produces stable reconstructions with low input variance, while RealNVP, MDN, and probabilistic cINN capture broader variability with slightly higher error.

On Burgers' equation, deterministic and invertible models fail to represent the inverse structure. Burgers' equation describes nonlinear wave dynamics with shock formation and diffusion, following the parabolic PDE $u_t + uu_x = \nu u_{xx}$. The inverse task is to recover the initial velocity profile from later-time observations.
The inverse map is strongly ill-posed: many distinct initial conditions evolve into nearly identical outputs.
Figure \ref{fig: classic pde results} shows that the deterministic and invertible models such as the MLP, INN, cINN, and iFNO all collapse to the mean of the input distribution. These models minimize re-simulation error but fail to capture variability in the input space. As a result, their predictions represent only a single plausible explanation for the observed data. The cVAE also collapses to the mean with a unimodal latent prior.
In contrast, RealNVP, MDN, and the probabilistic cINN generate diverse input samples that remain consistent under the forward operator, where all sampled inputs yield plausible reconstructions that match the observed outputs. 

Across these benchmarks, the behavior of probabilistic models depends on inverse structure. On near-bijective and mildly ill-posed inverse problems such as Darcy flow and elastic plate deformation, probabilistic models provide no performance advantage. They retain unnecessary variance and require additional training to suppress it in order to match deterministic accuracy. On the strongly ill-posed Burgers problem, probabilistic models are necessary to represent multiple plausible inputs consistent with the observed data. 

The invertible Fourier neural operator exhibits mixed behavior across the classical benchmarks. On Darcy flow, iFNO achieves the lowest re-simulation error, in part due to a substantially higher parameter count and longer training schedule than the B2B-based models. On elastic plate deformation, iFNO maintains low re-simulation error but produces noisier, less spatially coherent force fields, consistent with grid-level, pointwise inversion. These effects become more pronounced on Burgers' equation, where iFNO collapses to the conditional mean and produces noisy input predictions. Although the corresponding re-simulated outputs remain accurate, fine-scale structure in the input is distorted or lost. Overall, iFNO performs well on well-posed inverse problems but becomes increasingly sensitive to ambiguity and information loss as ill-posedness increases.

The classical PDE benchmarks show that inverse model performance is determined primarily by the structure of the inverse mapping. Deterministic and invertible coefficient-based models are sufficient when the inverse is well-posed or locally linear. As ill-posedness increases, deterministic inversion converges to conditional averages that obscure physically meaningful variability. Probabilistic inverse operators are therefore not uniformly beneficial, but are required when the inverse mapping is fundamentally non-unique.

\subsubsection{Performance on Inverse Operator Benchmarks}

\begin{figure}[!t]
    \centering
    \includegraphics[width=\linewidth]{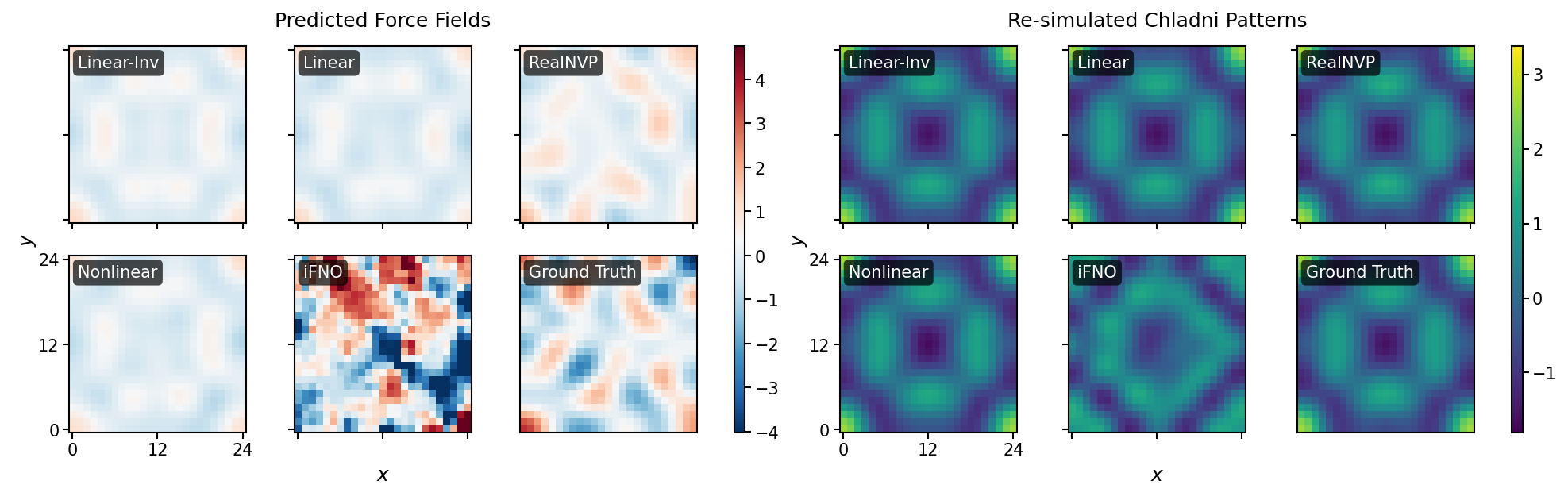}
    \includegraphics[width=\linewidth]{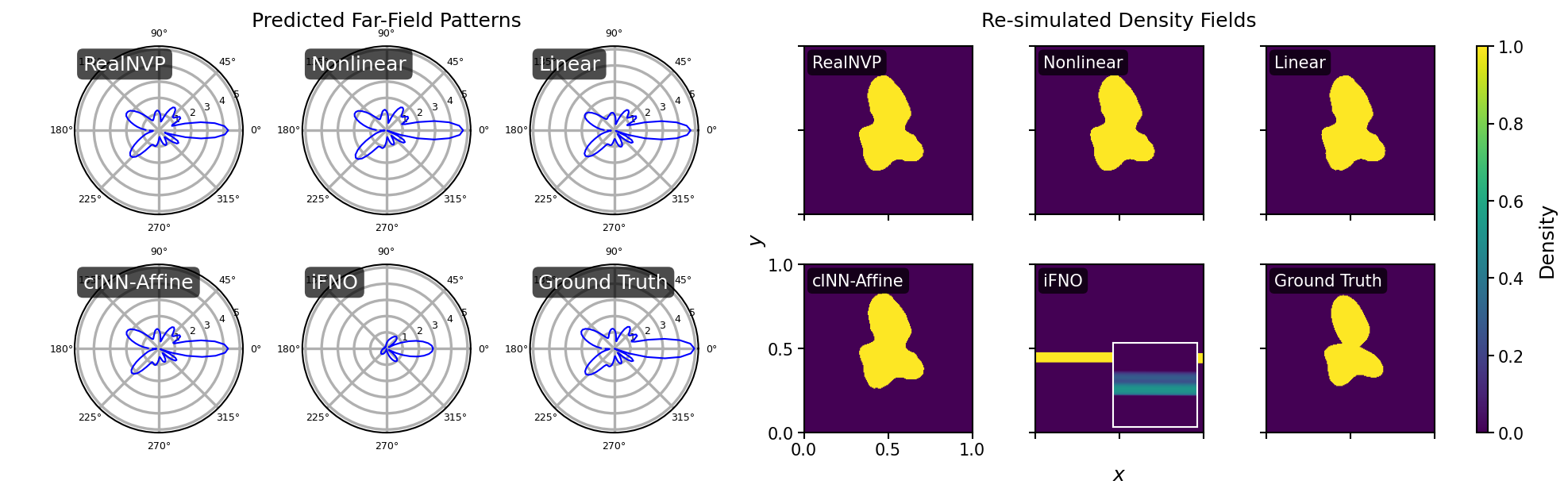}
    \caption{Qualitative comparison of inverse model performance on inverse operator benchmarks. We display representative test samples for the wave scattering and Chladni datasets, highlighting the top-performing models. For probabilistic models, input reconstructions are random samples from the predicted distribution, with corresponding output re-simulations shown on the right.}
    \label{fig: inverse results}
\end{figure}

We evaluate inverse neural operators on two novel inverse-operator benchmarks: Chladni plate resonance and wave scattering. Both tasks involve two-dimensional fields and forward operators with strong spectral characteristics, while the corresponding inverse mappings are highly ill-posed.

The Chladni plate resonance dataset models a fixed-time resonant response of a thin plate under harmonic excitation. 
Resonance patterns emerge from the eigenstructure of the system under fixed boundary conditions and strongly depend on the forcing frequency and the plate geometry. 
The goal is to recover input forcing modes from observed resonant vibration patterns on a thin elastic plate. 
Figure \ref{fig: inverse results} shows that the linear and nonlinear B2B models minimize re-simulation error but collapse to a trivial or mean forcing pattern. 
RealNVP and the probabilistic cINN produce varied patterns with comparable re-simulation accuracy. RealNVP, in particular, most closely matches the irregular forcing structures present in the dataset.

The wave scattering benchmark models the recovery of hidden 2D scatterer geometries from observed far-field wave measurements. We introduce this dataset to test inverse models on high-dimensional, spatially structured problems with multimodality and high-frequency content. 
Each sample consists of a complex-valued far-field pattern and its corresponding binary spatial density field, computed using a boundary integral solver for the Helmholtz equation. The far-field pattern reflects the angular scattering response of the shape. 
Figure \ref{fig: inverse results} shows that B2B-based models reconstruct the overall geometry and achieve low re-simulation error, but systematically miss fine-scale detail. Deterministic models recover mean shapes, while invertible and probabilistic models retain additional variability without degrading forward consistency. RealNVP and MDN, in particular, produce coherent, sample-consistent reconstructions that reflect plausible geometric variation.

By constraining predictions to a finite function space, B2B-based models produce smoother, globally coherent inputs and suppress high-frequency artifacts. In the Chladni problem, this smoothness bias is beneficial, as it regularizes the inversion. In the wave scattering problem, however, the same bias limits resolution, systematically suppressing sharp geometric boundaries even when they are present in the data. These results highlight an inherent tradeoff in the architecture. We anticipate that augmenting the neural basis architecture could improve the capture of fine-scale details, perhaps with frequency-based activations.

The invertible Fourier neural operator performs poorly in the inverse direction on both benchmarks. On the Chladni dataset, iFNO produces forcing patterns with sharp, spatially incoherent artifacts. On the wave scattering benchmark, its predictions frequently reduce to horizontal or vertical bands that fail to represent the underlying geometry. Although the corresponding re-simulated outputs capture dominant spectral modes, the fine-scale structure is distorted or lost. These behaviors are consistent with grid-level inversion that optimizes pointwise accuracy without enforcing global coherence, in contrast to coefficient-space inversion.

\subsubsection{Seismic Waveform Inversion}

\begin{figure}[!t]
    \centering
    \includegraphics[width=\linewidth]{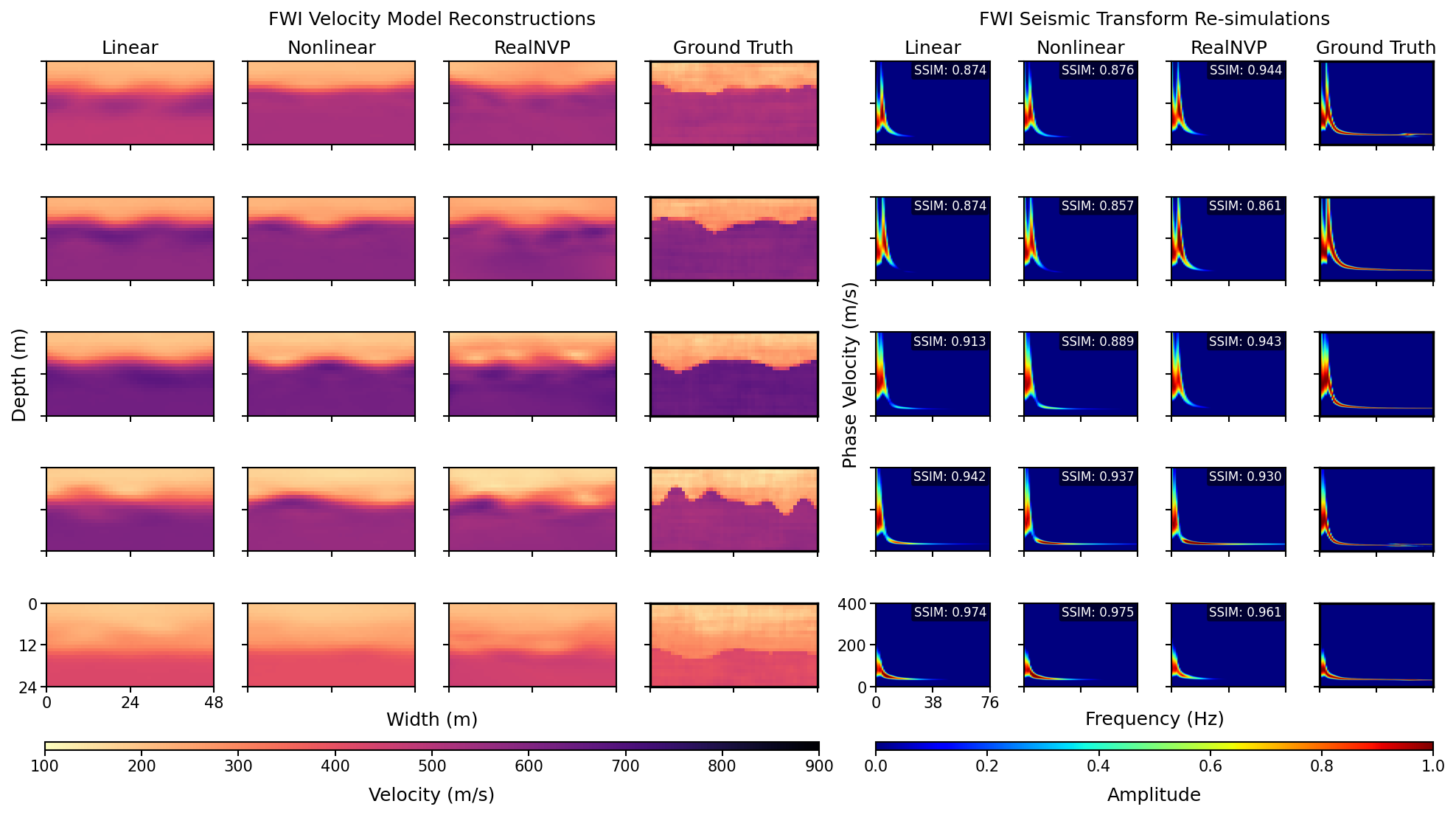}
    \caption{Qualitative comparison of inverse models on the FWI benchmark. We show representative reconstructions of subsurface velocity models (left) and corresponding seismic re-simulations (right) for the top-performing inverse models. While models generally recover large-scale structural features of the velocity field, they often fail to capture high-frequency components, resulting in smoothed re-simulated wavefields.}
    \label{fig: fwi results}
\end{figure}

We evaluate the proposed inverse operator framework on a full waveform inversion (FWI) benchmark for near-surface seismic imaging introduced in \cite{ABBAS2023105305, 10.1093/gji/ggac179}. The dataset consists of paired two-dimensional subsurface velocity profiles and frequency-velocity representations of surface-wave measurements, following the synthetic setup introduced in \cite{ABBAS2023105305}. 
Full waveform inversion is a canonical inverse problem in geophysics that seeks to recover subsurface velocity models from seismic wavefield observations. FWI is highly ill-posed and nonlinear, where multiple subsurface profiles can explain the same wavefield. 

At a macro level, Figure \ref{fig: fwi results} shows that linear and nonlinear deterministic inverse models produce accurate subsurface velocity predictions that capture the dominant interface. 
Across models, the predicted subsurface profiles recover the large-scale structure of the velocity field, including the overall layering and the location of the primary interface.
The probabilistic RealNVP model predicts subsurface velocity profiles that are less accurate when compared to ground truth than the deterministic models, though its forward re-simulations have finer spectral detail. 

Fine-scale structure is smoothed in both the predicted subsurface profiles and the corresponding frequency responses. Sharp boundaries between layers are attenuated in the inverse predictions, and high-frequency components are suppressed in the forward re-simulations. These effects are visible in the figure as blurred interfaces in the velocity profiles and reduced resolution in the frequency-velocity domain. 
The observed smoothing arises from two identifiable sources. First, the learned coefficient basis favors smooth velocity fields by construction, which stabilizes inversion but suppresses sharp interfaces between subsurface layers. Second, the learned forward operator further smooths high-frequency content when propagating predicted subsurface models to the frequency domain. 

Quantitatively, most inverse models achieve relative $L^2$ re-simulation errors on the order of $10^{-1}$ across the FWI dataset, as shown in Table \ref{tab: results}. This error level is consistent across model classes and reflects a limitation of the evaluation metric rather than inverse performance. Specifically, while the function encoders for the input and output spaces achieve reconstruction errors on the order of $10^{-2}$, the learned nonlinear forward model converges to errors on the order of $10^{-1}$. Because inverse predictions are evaluated through this learned forward operator, the reported error reflects forward consistency under a surrogate model rather than direct inverse reconstruction accuracy.

Overall, the full waveform inversion results show that the proposed inverse operator framework functions as a stable surrogate for recovering coarse subsurface structure without imposing problem- or instance-specific priors. The approach consistently produces coherent velocity profiles and forward-consistent predictions, but does not resolve sharp boundaries or high-frequency detail. Improving resolution in this regime will require incorporating additional prior information, such as learning residual corrections relative to a coarse baseline model, augmenting the neural basis architectures to capture finer detail, or constraining the inverse to an affine subspace. 

\subsection{Analysis of Stability and Sensitivity}

\begin{figure}
    \centering
    \includegraphics[keepaspectratio]{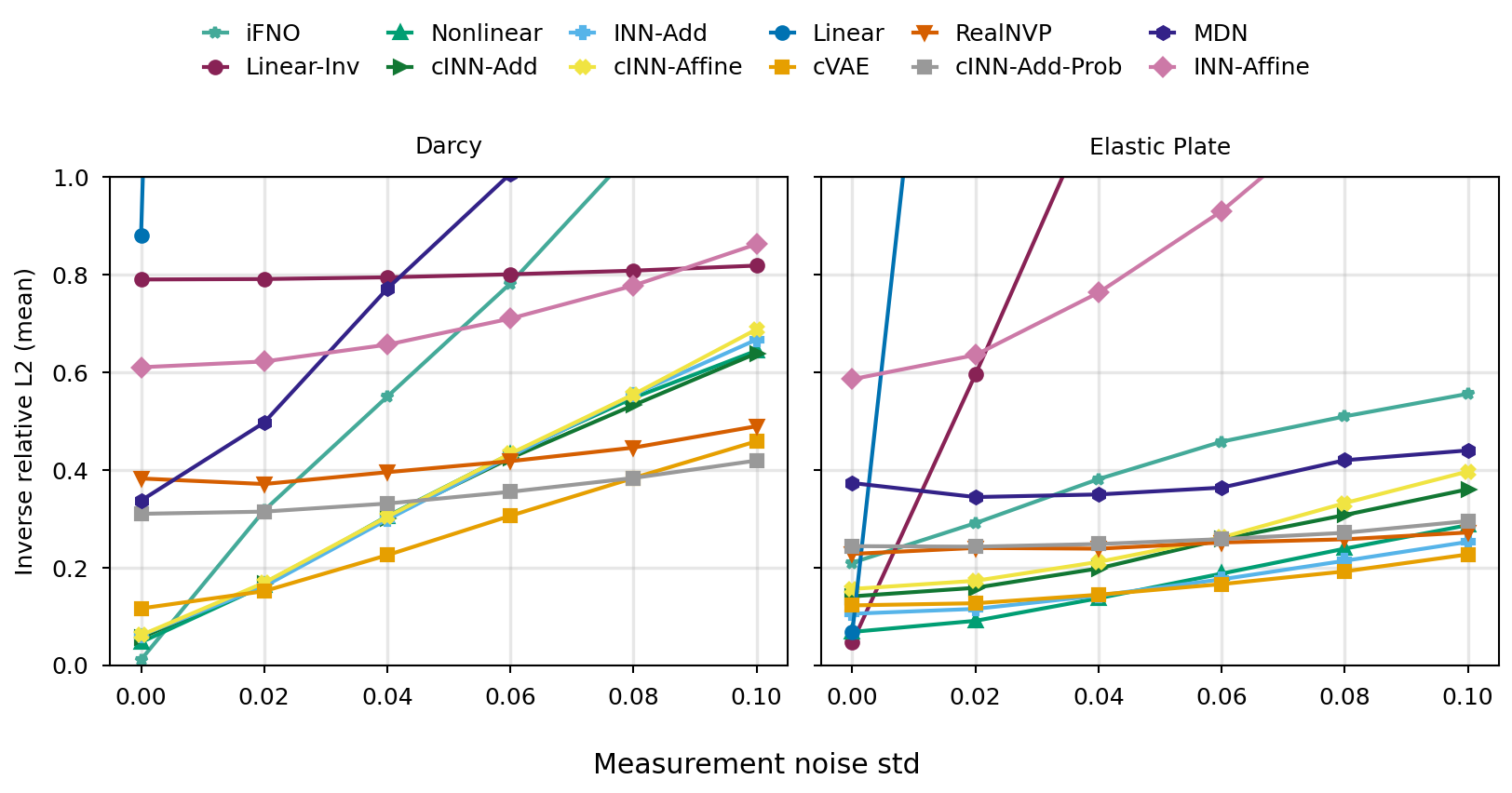}
    \caption{Sensitivity of inverse models to observation noise on the elastic plate benchmark.}
    \label{fig: analysis}
\end{figure}

We next study how measurement noise affects the stability of learned inverse operators. All six benchmarks considered so far use noise-free data, which favors accuracy but does not reflect realistic settings, which often involve noisy or incomplete measurements. To test robustness, we add increasing levels of noise to the output measurements and track how inverse predictions degrade.

We perform this analysis on the Darcy flow and elastic plate deformation benchmarks, where we can directly compare predicted inputs to ground truth. We corrupt output measurements with zero-mean Gaussian noise with standard deviation ranging from $0$ to $0.1$. A noise level of $0.1$ represents strong corruption relative to the signal and visibly degrades the measurements. We measure stability using relative $L_2$ error between predicted and true inputs.

Deterministic and invertible models lose accuracy rapidly as noise increases. Figure \ref{fig: analysis} shows that linear, nonlinear, and invertible models exhibit a significant increase in error as noise variance grows. The nonlinear MLP, which achieves one of the lowest errors on noise-free data on both datasets, degrades sharply, likely because it relies on fine-grained correlations that amplify perturbations. The linear models degrade rapidly, as expected, quickly losing fidelity since the linear operator amplifies noise in the coefficient estimate. 

Importantly, probabilistic models behave differently and remain stable across noise levels. RealNVP and probabilistic cINN variants maintain nearly constant error as noise increases, while the cVAE shows a moderate but controlled degradation. 
Unlike deterministic and invertible models, these approaches do not learn a single pointwise inverse mapping from outputs to inputs. Instead, they learn a conditional distribution over plausible input coefficients given the observed output coefficients. Because the learned inverse is distributional rather than one-to-one, small perturbations in the measurements do not force large changes in the predicted input distribution. The models can adjust the probability mass within the learned distribution without committing to a different deterministic solution. As a result, probabilistic models trade accuracy in the noise-free setting for robustness under perturbations, leading to slower degradation as noise increases.

The iFNO baseline degrades more rapidly than other deterministic models. iFNO operates directly on grid-based functional data, which allows pointwise errors or noise to propagate through the inverse map. In contrast, B2B-based models first project functional data onto a learned coefficient basis via regularized least squares. This projection acts as an explicit de-noising step and improves stability across both benchmarks.

This analysis exposes a clear tradeoff between accuracy and robustness. Deterministic and invertible models perform best on noise-free data but degrade as measurement noise increases. Probabilistic inverse operators lose some accuracy in the noise-free setting but gain stability, making them better suited to noisy or incomplete measurement settings. These results reinforce the need to select inverse model classes based on both inverse structure and data quality.

\section{Discussion}

We propose a principled formulation of inverse neural operators that separates function representation from operator learning through basis-to-basis (B2B) mappings. By making the function representation explicit and performing inversion in learned coefficient spaces, this approach avoids fixed discretizations and supports deterministic, invertible, and probabilistic inverse models within a single framework.
Our core innovation lies in architectures that model inverse maps between learned function space coefficient representations, capturing uncertainty and non-uniqueness while enabling interpretability and modularity.

Our results demonstrate a clear benefit of our architecture, enabling different architectural choices, enabling probabilistic neural operators, and also showing that we can capture ill-posedness and uncertainty.
In contrast to operator-learning methods such as DeepONet \cite{lu2021learning}, Fourier Neural Operators (FNO) \cite{li2021fourier}, and their variants \cite{10.1145/3648506, goswami2023physics, kontolati2024learning}, which entangle representation and operator learning within a single network, B2B makes the representation explicit. 
Reducing the operator problem to learning a coefficient mapping between $\alpha$ and $\beta$ makes the representation interpretable, enables direct recovery of inputs from outputs, and provides a flexible framework for inverse operators.

Despite the promising performance modeling inverse PDE problems, our proposed approach can be further improved. 
In particular, the learned function bases impose smoothness that stabilizes inversion but limits resolution in problems with sharp interfaces or non-smooth behavior. This bias stabilizes inversion in ambiguous settings but limits resolution in problems that require sharp interfaces or high-frequency detail, such as wave scattering and full waveform inversion. More expressive or adaptive basis architectures, including frequency-aware representations or multiscale constructions, could reduce this bias.
Another challenge lies in integrating prior physics knowledge. Incorporating prior information or physics, for example through residual inversion relative to a coarse baseline or by restricting inversion to an affine subspace, offers another practical path to improving resolution without sacrificing robustness.
Finally, the B2B decomposition creates a natural bridge to kernel methods and reproducing kernel Hilbert space theory \cite{scholkopf2002learning, low2025functionspaceskernelslearning}, which may enable tighter analysis, adaptive basis design, and principled error control.
These potential improvements highlight that our approach is not tied to a specific architecture. Rather, it offers a flexible framework that can evolve with advances in representation learning and probabilistic modeling.

\section{Conclusion}

In this work, we present a principled approach to learning inverse neural operators that separates representation from operator learning. This separation lets us build generalizable, invertible, and probabilistic inverse maps without fixed discretizations. 
Our proposed method combines basis-to-basis representations with deterministic and probabilistic architectures, giving a simple and flexible tool for ill-posed inverse problems. 

The broader impact of this work lies in its potential to enable scalable surrogate solutions for inverse problems. 
Inverse problems are central to many areas of engineering, including nondestructive testing and thermal imaging \cite{isakov2006inverse}, to seismic inversion \cite{10.1190/1.1441754, o2023imaging}, materials design \cite{jmi.2021.07}, and robotics \cite{10.1145/1015330.1015430}. 
In many of these settings, fast and reliable inverse surrogates are more practically useful for real-time decision-making than exact reconstruction.
Our approach provides a path toward scalable and principled solutions for surrogate modeling, with the potential to accelerate progress in multiple domains.

\section{Methods}

We now describe the proposed inverse operator learning framework. 
We first formalize the inverse operator learning problem as a mapping from output functions to distributions over input functions. We then introduce a basis-to-basis representation that maps functions in the input and output spaces to finite-dimensional coefficient vectors, enabling inversion to be performed in a learned coefficient space. Finally, we describe the inverse model classes considered, including deterministic, invertible, and probabilistic architectures, along with the training and evaluation procedures. 
Additional architectural and implementation details are provided in the Supplementary Methods.

\subsection{Mathematical Formulation}

Let $\mathcal{G}$ and $\mathcal{H}$ be Hilbert spaces of functions. Let $T : \mathcal{G} \to \mathcal{H}$ be the forward operator between $\mathcal{G}$ and $\mathcal{H}$. 
We formulate inverse operator learning as learning a map $\mathcal{A}_{\theta} : \mathcal{H} \to \mathcal{P}(\mathcal{G})$ that outputs a probability distribution $p \in \mathcal{P}(\mathcal{G})$ over plausible input function reconstructions. Rather than predicting a single reconstruction, the inverse operator $\mathcal{A}_{\theta}$ predicts a distribution over plausible inputs $f \in \mathcal{G}$ consistent with measurements from an output function $h \in \mathcal{H}$.

We assume access to a forward model $T$ (or a learned forward model) during training and inference. 
The training data consists of a set of $N \in \mathbb{N}$ datasets $\lbrace D_{1}, \ldots, D_{N} \rbrace$. Each dataset $D_{n}$, $n = 1, \ldots, N$, consists of input-output pairs for both the input and output function spaces, $D_{n} = \lbrace \lbrace (x_{i}, f_{n}(x_{i})) \rbrace_{i=1}^{m}, \lbrace (y_{i}, h_{n}(y_{i})) \rbrace_{i=1}^{p} \rbrace$, where $h_{n} = T(f_{n})$ is the output of the forward operator.
Note that the input and output coordinates $\lbrace x_{i} \rbrace$ and $\lbrace y_{i} \rbrace$ may differ across samples and do not need to lie on a fixed grid or discretization. 
After training, we only have access to measurements $\lbrace (y_{i}, h(y_{i})) \rbrace_{i=1}^{q}$ from a new, unseen output function $h$ that we want to evaluate. 

We model measurement noise as additive noise on output observations. We generate noisy measurements as $h(y_i) = (T f)(y_i) + \epsilon_i$, where $\epsilon_i \sim \mathcal{N}(0,\sigma^2)$ independently across measurement locations. We train inverse models on noise-free observations and evaluate robustness by injecting noise at test time with specified $\sigma$ values

\subsection{Basis-to-Basis Representation Learning}

We follow a two-stage approach to inverse operator learning: first, we learn neural basis function representations for input and output function spaces; then, we learn the inverse mapping between their coefficient representations. 

Our approach builds on basis-to-basis learning \cite{INGEBRAND2025117646}, which represents functions as linear combinations of learned basis functions parameterized by neural networks. 
We learn neural network basis functions $\lbrace \psi_{1}, \ldots, \psi_{k} \rbrace$ and $\lbrace \varphi_{1}, \ldots, \varphi_{l} \rbrace$ to represent functions in $\mathcal{G}$ and $\mathcal{H}$, respectively \cite{pmlr-v267-ingebrand25a}. 
Each basis function is implemented as a feed-forward neural network. 
Then, functions can be represented as linear combinations of the learned basis functions,
\begin{equation}
    \hat{f}(x) = \sum_{j=1}^{k} \alpha_{j} \psi_{j}(x; \theta), \quad \hat{h}(y) = \sum_{j=1}^{l} \beta_{j} \varphi_{j}(y; \theta),
\end{equation}
where $\theta$ are the learned neural network parameters.
For any function $f$ and data $\lbrace(x_{i}, f(x_{i})) \rbrace_{i=1}^{m}$, we compute the corresponding coefficients $\alpha$ as the solution to a least squares problem, which admits a closed-form solution via the normal equations $(G + \lambda I) \alpha = F$, where $G_{ij} = \langle \psi_{i}, \psi_{j} \rangle$ and $F_{i} = \langle \psi_{i}, f \rangle$. 
The inner product is computed using Monte Carlo integration over the function domain.
We train the basis functions by minimizing the reconstruction error across all training functions, 
\begin{equation}
    \min_{\theta} \frac{1}{Nm} \sum_{n=1}^{N} \sum_{i=1}^{m} \lVert f_{n}(x_{i}) - \hat{f}_{n}(x_{i}) \rVert^{2} + \lambda \lVert \alpha^{n} \rVert^{2},
\end{equation}
where $\lambda > 0$ is the regularization parameter. We use $\lambda = 10^{-3}$ in our experiments.
We use an identical procedure to compute the coefficients $\beta$ for a function $h$ in the output function space and to train the basis functions $\lbrace \varphi_{1}, \ldots, \varphi_{l} \rbrace$.
Prior work in function encoders provides theoretical analysis of approximation error and representation capacity once the basis functions are fixed \cite{INGEBRAND2025117646, low2025functionspaceskernelslearning}.
We refer the reader to \cite{INGEBRAND2025117646, pmlr-v267-ingebrand25a} for the derivation of the basis function representation, regularized least squares coefficient optimization, and training procedure.

\subsection{Inverse Architectures}

We perform inverse operator learning by mapping output coefficients $\beta \in \mathbb{R}^{l}$ to input coefficients $\alpha \in \mathbb{R}^{k}$. We adapt standard deterministic, invertible, and probabilistic architectures by replacing function-valued inputs and outputs with coefficient vectors.
Additional model details are provided in Supplementary Methods, Section \ref{section: supplementary methods inverse models}.

\subsubsection{Deterministic Models}

We consider three deterministic inverse models: a linear regressor, a linear model computed via the forward map, and a nonlinear (MLP) map. These models assume that the inverse mapping is unique and stable in coefficient space.

\paragraph{Linear.}
The linear model learns a matrix $W \in \mathbb{R}^{k \times l}$ by solving a least-squares regression problem that maps $\beta$ directly to $\alpha$. In contrast, the linear-inverse model first learns the forward linear map from $\alpha$ to $\beta$ and then computes its inverse when the matrix is square and well-conditioned. 

\paragraph{MLP.}
The nonlinear MLP directly learns a parametric map $\beta \mapsto \alpha$ using mean-squared error loss. 

\subsubsection{Invertible Models}

We implement invertible neural networks (INNs) and conditional invertible neural networks (cINNs) to model deterministic, bijective relations between coefficient vectors. These models assume that the inverse mapping is approximately bijective in coefficient space and rely on architectural constraints rather than sampling to perform inversion. 

\paragraph{INN.}
The standard INNs learn an invertible map between input and output coefficients, mapping $\alpha \mapsto \beta$ in the forward direction and $\beta \mapsto \alpha$ through exact architectural inversion. These models use coupling-based architectures that enforce invertibility by construction.

\paragraph{cINN.}
The conditional variants (cINNs) augment this architecture by introducing a latent variable and conditioning both the forward and inverse coupling transformations on the output coefficients $\beta$. In this setting, the latent variable does not represent a learned probability distribution. We fix the latent variable to zero at inference time and treat the model as a deterministic inverse map conditioned on $\beta$. We evaluate both additive and affine coupling layers in this deterministic setting. 

\subsubsection{Probabilistic Models}

We model non-unique inverse mappings by learning a map from $\beta$ to distributions $p(\alpha)$ in coefficient space. We consider four probabilistic architectures: a conditional variational autoencoder (cVAE), a mixture density network (MDN), a conditional normalizing flow (RealNVP), and a probabilistic conditional invertible neural network with additive coupling layers (cINN-Additive-Prob).
All probabilistic models generate inverse predictions by sampling coefficient vectors $\alpha$ from the learned conditional distribution and reconstructing input functions using the learned input basis.

\paragraph{cVAE.}
The cVAE models $p(\alpha \mid \beta)$ using a latent variable formulation. During training, the encoder maps the joint coefficient pair $(\alpha, \beta)$ to a latent variable z by learning an approximate posterior $q(z \mid \alpha, \beta)$. The decoder conditions on $\beta$ and a sampled latent variable $z$ to reconstruct $\alpha$. The model trains by optimizing the standard evidence lower bound, which balances reconstruction accuracy with regularization of the latent distribution. At inference time, the model samples $z$ from the prior and decodes samples of $\alpha$ conditioned on $\beta$.

\paragraph{MDN.}
The MDN directly parameterizes the conditional distribution $p(\alpha \mid \beta)$ as a finite mixture of Gaussian components. Given $\beta$, the network outputs mixture weights, means, and diagonal covariances that define the distribution over $\alpha$. The model trains by maximizing the conditional log-likelihood of the observed coefficients. At inference time, the model samples $\alpha$ by first sampling a mixture component and then sampling from the corresponding Gaussian.

\paragraph{RealNVP.}
The conditional normalizing flow defines an invertible transformation between input coefficients $\alpha$ and a latent variable $z \sim \mathcal{N}(0, I)$, conditioned on $\beta$. The model uses a sequence of masked affine coupling layers, where each transformation conditions on $\beta$. Training maximizes the exact conditional log-likelihood using the change-of-variables formula. At inference time, the model samples $z$ from the Gaussian base distribution and applies the inverse transformation to generate samples of $\alpha$.

\paragraph{cINN (Probabilistic).}
The probabilistic cINN adapts coupling-based invertible architectures for conditional density estimation. The model uses additive coupling layers conditioned on $\beta$ and introduces stochasticity by reparameterizing a latent variable during training. Unlike the deterministic cINN, this variant explicitly models a conditional distribution over $\alpha$ and trains by maximizing the conditional log-likelihood. At inference time, the model samples the latent variable to generate diverse coefficient reconstructions. We restrict this probabilistic variant to additive coupling layers, as the affine formulation closely parallels the RealNVP architecture.

\subsection{Training Setup}

We use a common training configuration across all datasets unless otherwise noted. 
We standardize output sampling locations across models within each dataset to ensure direct comparability across architectures and baselines.
We train separate function encoders for the input and output spaces using $k=l=100$ basis functions. 
We select this value as a conservative overestimate based on prior basis-to-basis operator learning results \cite{INGEBRAND2025117646}. Overparameterizing the basis does not degrade performance, as redundant basis functions collapse during training, while underparameterization limits expressivity. 
In principle, one can estimate a minimal basis dimension for a given dataset using intrinsic dimension estimates such as PCA. 
However, this requires dataset-specific information that is generally not available at inference time.

Each basis function is parameterized by a multilayer perceptron with three hidden layers of width 512 and ReLU activations. We choose this architecture as a stable middle ground between expressivity and optimization cost across all benchmarks. We train the function encoders and the forward basis-to-basis operator using mini-batches of 20 functions per batch for 20,000 gradient steps with a learning rate of $10^{-4}$.

We train all inverse models on fixed coefficient representations produced by the trained function encoders. To control for model capacity, we match the parameter count of all inverse architectures to that of a three-layer MLP with width 512. Deterministic inverse models train using mean-squared error loss. Probabilistic models train using the corresponding conditional likelihood or variational objective described in the previous section.

For comparison, we evaluate the invertible Fourier neural operator (iFNO) \cite{pmlr-v258-long25a}. iFNO operates directly on discretized function values rather than coefficient representations and uses a three-stage training procedure. The architecture consists of three learnable components:  
\begin{enumerate*}[label=(\roman*)]
\item 
lifting and projection networks that map between physical coordinates and a latent channel space, 
\item 
an invertible Fourier backbone composed of coupling blocks with spectral convolution layers, and 
\item 
a dataset-matched variational autoencoder that regularizes the inverse mapping and enables posterior inference.
\end{enumerate*}
Across datasets, the invertible backbone uses three coupling blocks, each with two Fourier transformations, resulting in six Fourier layers interleaved with pointwise multilayer perceptrons and skip connections. The VAE architecture adapts to the spatial dimensionality of each dataset and includes multiple convolutional encoder and decoder layers with a low-dimensional latent space. Depending on dataset resolution and geometry, the total parameter count ranges from approximately $1.4\times 10^6$ to $5.1\times 10^6$ parameters. We train iFNO using the authors' recommended hyperparameters and report results from the best-performing configurations. 

\subsubsection{Training and Re-simulation Evaluation}
\label{section: resimulation error}

To evaluate and compare the inverse models, we compute a \emph{re-simulation loss} based on a separately learned forward model. 
We train the forward operator $\hat{T}$ using the B2B framework to map input coefficients $\alpha$ to output coefficients $\beta$, using a multilayer perceptron trained to minimize MSE between predicted and true coefficients over mini-batches, following the training procedure in \cite{INGEBRAND2025117646}. We fix this forward model during all inverse evaluations.
We use re-simulation loss because direct evaluation in input space is ill-posed and divergence estimation is intractable.
Inverse problems are often non-unique, so comparing predictions in input space $\mathcal{G}$ can be misleading without access to the full conditional distribution $p(f \mid h)$, which is unknown. Moreover, divergence metrics like KL divergence, Wasserstein distance, or MMD cannot be computed because we observe only one input-output pair per sample---not multiple $f$ values per $h$. Even if such samples were available, estimating high-dimensional function distributions requires impractically large sample sizes. Re-simulation loss sidesteps these challenges by testing whether predicted inputs reproduce the observed output under the forward model, which is the goal of inverse inference.

For each predicted input $\hat{f} \sim \mathcal{A}_{\theta}(h)$, we compute the corresponding output $\hat{h} = \hat{T}(\hat{f})$ using the learned forward model $\hat{T}$, and evaluate the relative $L_2$ error $\lVert h - \hat{h} \rVert / \lVert h \rVert$. We evaluate the re-simulation error in function space at the observed output measurement locations $\lbrace y_i \rbrace$, rather than in coefficient space, to assess consistency with the measured data.
For deterministic models, this is a single prediction. For probabilistic models, we estimate the expected loss over samples:
\begin{equation}
    \label{eqn: resimulation loss}
    \mathbb{E}_{\hat{f} \sim \mathcal{A}_{\theta}(h)} \biggl[ \frac{\lVert h - \hat{T}(\hat{f}) \rVert}{\lVert h \rVert} \biggr].
\end{equation}
We approximate this expectation using 10 independent Monte Carlo samples per test instance.
Critically, because re-simulation relies on a learned forward operator, reported errors reflect both inverse-model error and approximation error in the forward surrogate. In particular, a perfect inverse may still incur non-zero re-simulation error due to imperfections in the forward surrogate.

For the iFNO baseline, we compute re-simulation error by applying the learned inverse map to obtain an input reconstruction and then propagating this reconstruction through the learned forward map, not the B2B surrogate. We evaluate the resulting output at the same measurement locations $\lbrace y_i \rbrace$, ensuring that re-simulation error is computed in observation space for all models.

We report the mean relative $L_2$ re-simulation error in Table \ref{tab: results}. To complement this, we also report the MSE between observed and re-simulated outputs in Supplementary Table \ref{tab: results mse}.

\bibliographystyle{plain}
\bibliography{bibliography}

\section{Acknowledgments}

This work was funded in part by AFOSR FA9550-19-1-0005 and by NSF 2214939, 2409535, 2436738, 2339678, 2321040 and 2438193. Any opinions, findings, conclusions, or recommendations expressed in this material are those of the author(s) and do not necessarily reflect the views of the funding organizations.

We would like to thank Somdatta Goswami for helpful discussion and feedback.
We also thank Joseph Vantassel, Brady Cox and Aser Abbas for sharing the FWI dataset used in this paper.

\appendix
\section{Results}

\begin{table*}[!ht]
\scriptsize
\centering
\begin{tabular*}{\textwidth}{lcccccc}
\toprule
 & Burgers & Darcy Flow & Chladni & Elastic & Wave Scattering & FWI \\
\midrule
\multicolumn{7}{l}{\textbf{Inverse Model Re-simulation MSE}} \\
Linear
  & $6.76 \pm 10.63 \mathrm{e}^{-1}$
  & $1.91 \pm 1.16 \mathrm{e}^{-1}$
  & $2.79 \pm 2.16 \mathrm{e}^{-3}$
  & $8.06 \pm 15.95 \mathrm{e}^{-1}$
  & $1.08 \pm 0.73 \mathrm{e}^{-1}$
  & $2.89 \pm 0.16 \mathrm{e}^{-2}$ \\
Linear-Inv
  & $3.03 \pm 2.56$
  & $5.57 \pm 6.64 \mathrm{e}^{1}$
  & $5.98 \pm 11.92\mathrm{e}^{7}$
  & $5.43 \pm 6.23 \mathrm{e}^{-1}$
  & $8.14 \pm 10.22 \mathrm{e}^{1}$
  & $1.94 \pm 3.03 e^+4$ \\
MLP
  & $2.48 \pm 1.27 \mathrm{e}^{-3}$
  & $3.52 \pm 2.58 \mathrm{e}^{-3}$
  & $6.48 \pm 2.30 \mathrm{e}^{-3}$
  & $1.77 \pm 0.62 \mathrm{e}^{-3}$
  & $7.05 \pm 0.90 \mathrm{e}^{-2}$
  & $3.75 \pm 0.18 \mathrm{e}^{-2}$ \\
\midrule
INN-Additive
  & $3.13 \pm 2.98 \mathrm{e}^{-2}$
  & $2.55 \pm 1.95 \mathrm{e}^{-2}$
  & $1.20 \pm 0.14 \mathrm{e}^{-2}$
  & $1.52 \pm 1.23 \mathrm{e}^{-2}$
  & $9.05 \pm 1.04 \mathrm{e}^{-2}$
  & $8.43 \pm 1.48 \mathrm{e}^{-2}$ \\
cINN-Additive
  & $3.67 \pm 2.69 \mathrm{e}^{-3}$
  & $1.10 \pm 0.62 \mathrm{e}^{-2}$
  & $6.97 \pm 2.18 \mathrm{e}^{-3}$
  & $2.56 \pm 0.90 \mathrm{e}^{-3}$
  & $7.06 \pm 0.91 \mathrm{e}^{-2}$
  & $5.92 \pm 0.93 \mathrm{e}^{-2}$ \\
INN-Affine
  & $7.17 \pm 4.51 \mathrm{e}^{-1}$
  & $3.38 \pm 2.37$
  & $6.35 \pm 0.47 \mathrm{e}^{-1}$
  & $1.01 \pm 0.46$
  & $4.27 \pm 0.09 \mathrm{e}^{-1}$
  & $2.68 \pm 0.44 \mathrm{e}^{-1}$ \\
cINN-Affine
  & $2.78 \pm 1.50 \mathrm{e}^{-3}$
  & $6.98 \pm 1.90 \mathrm{e}^{-3}$
  & $1.13 \pm 0.18 \mathrm{e}^{-2}$
  & $4.45 \pm 1.34 \mathrm{e}^{-3}$
  & $7.10 \pm 0.93 \mathrm{e}^{-2}$
  & $6.09 \pm 0.85 \mathrm{e}^{-2}$ \\
\midrule
cVAE
  & $6.73 \pm 1.51 \mathrm{e}^{-3}$
  & $9.69 \pm 2.24 \mathrm{e}^{-3}$
  & $9.39 \pm 2.30 \mathrm{e}^{-3}$
  & $2.95 \pm 0.48 \mathrm{e}^{-3}$
  & $7.20 \pm 0.90 \mathrm{e}^{-2}$
  & $3.82 \pm 0.28 \mathrm{e}^{-2}$ \\
MDN
  & $1.11 \pm 0.17 \mathrm{e}^{-1}$
  & $1.33 \pm 0.62 \mathrm{e}^{-1}$
  & $2.56 \pm 0.10 \mathrm{e}^{-1}$
  & $8.35 \pm 4.12 \mathrm{e}^{-3}$
  & $7.28 \pm 0.88 \mathrm{e}^{-2}$
  & $6.41 \pm 1.21 \mathrm{e}^{-1}$ \\
cINN-Additive-Prob 
  & $1.24 \pm 0.64 \mathrm{e}^{-2}$
  & $7.13 \pm 0.85 \mathrm{e}^{-2}$
  & $8.74 \pm 2.49 \mathrm{e}^{-3}$
  & $5.72 \pm 2.10 \mathrm{e}^{-3}$
  & $7.14 \pm 0.81 \mathrm{e}^{-2}$
  & $7.17 \pm 0.42 \mathrm{e}^{-2}$ \\
RealNVP
  & $8.83 \pm 6.95 \mathrm{e}^{-3}$
  & $2.56 \pm 0.93 \mathrm{e}^{-2}$
  & $2.38 \pm 0.36 \mathrm{e}^{-2}$
  & $2.27 \pm 2.22 \mathrm{e}^{-2}$
  & $7.12 \pm 0.85 \mathrm{e}^{-2}$
  & $5.52 \pm 0.44 \mathrm{e}^{-2}$ \\
\midrule
iFNO
  & $5.42 \pm 2.10 \mathrm{e}^{-3}$ 
  & $7.54 \pm 2.00 \mathrm{e}^{-6}$ 
  & -- 
  & $4.62 \pm 0.32 \mathrm{e}^{-1}$ 
  & $8.46 \pm 0.05 \mathrm{e}^{-1}$ 
  & -- \\
\bottomrule
\end{tabular*}
\caption{Mean squared error of re-simulated outputs across benchmark PDE inverse problems.}
\label{tab: results mse}
\end{table*}

\subsection{Benchmark Details}
\label{section: supplementary results benchmarks}

Across all benchmark datasets, we consider forward operators that map an input function $f$ defined on a spatial domain to an output function $h = T(f)$ obtained by solving a governing partial differential equation. Input functions are sampled from problem-specific distributions and propagated through the corresponding forward solver to generate paired input–output data.

For all datasets, input and output functions are evaluated at fixed sets of spatial locations that are held constant across samples. Observation noise is not added unless otherwise stated. Inverse learning is performed using only the sampled output functions, with access to the full forward model limited to training.

\subsection{Darcy Flow}


This benchmark considers a nonlinear one-dimensional Darcy-type forward operator and its associated inverse task, adapted from \cite{INGEBRAND2025117646}. The forward operator maps an input field to the solution of a nonlinear elliptic equation.

The forward problem is defined on $x \in [0,1]$ by
\begin{equation}
-\frac{d}{dx}\biggl(\kappa(s(x))\,\frac{ds(x)}{dx}\biggr) = u(x),
\end{equation}
with homogeneous Dirichlet boundary conditions $s(0)=s(1)=0$. The permeability depends on the solution through
$\kappa(s(x)) = 0.2 + s^2(x)$.

The inverse task is to recover the input field $u(x)$ from observations of the solution $s(x)$. For this class of nonlinear elliptic equations, the forward operator is smooth and monotone under the specified boundary conditions. As a result, when interior solution observations are available, the inverse mapping from solution to input is well-posed and effectively bijective.

The input field $u(x)$ is drawn from a zero-mean Gaussian random field with squared-exponential covariance
\begin{equation}
k(x,x') = \exp\biggl(-\frac{\|x-x'\|^2}{2\sigma^2}\biggr),
\end{equation}
with length scale $\sigma = 0.04$. For each realization, the forward problem is solved using a finite-element discretization, following the procedure described in \cite{INGEBRAND2025117646}.

Both the input field $u(x)$ and the solution $s(x)$ are evaluated at fixed sets of spatial locations, which are held constant across all samples.

\subsection{Burger's Equation}


This benchmark considers a nonlinear time-dependent forward operator governed by Burgers' equation, adapted from \cite{INGEBRAND2025117646}. The forward task maps an initial velocity profile to a velocity field at a later time.

The forward dynamics are defined by the viscous Burgers' equation
\begin{equation}
\frac{\partial u}{\partial t}(x,t)
=
\nu \frac{\partial^2 u}{\partial x^2}(x,t)
-
u(x,t)\frac{\partial u}{\partial x}(x,t),
\end{equation}
with initial condition $u(x,0)=f(x)$ and periodic boundary conditions. The viscosity parameter $\nu>0$ controls the strength of dissipation.

The inverse task is to recover the initial condition $f(x)$ from observations of the solution at a fixed final time. In contrast to the Darcy problem, this inverse mapping is intrinsically ill-posed. Dissipation smooths high-frequency components of the initial condition, and nonlinear advection can merge characteristics, causing distinct initial states to evolve into similar final-time solutions. This information loss occurs even in the absence of observation noise and induces non-uniqueness in the inverse problem.

Initial conditions are drawn from a zero-mean Gaussian random field with covariance operator $(-\Delta + 5^2 I)^{-4}$, following the source dataset. The resulting final-time velocity fields constitute the output functions used for inverse learning.

\subsection{Elastic Plate}

This benchmark considers a two-dimensional linear elasticity problem under plane stress conditions, adapted from prior operator learning work. The forward operator maps applied boundary forces to displacement fields in an elastic plate.

The forward problem is defined on a thin rectangular plate by the equilibrium equations
\begin{equation}
\nabla \cdot \sigma + \boldsymbol{f}(\boldsymbol{x}) = 0,
\end{equation}
where $\sigma$ denotes the Cauchy stress tensor and $\boldsymbol{f}(\boldsymbol{x})$ represents an applied force field. In our dataset, the forcing is applied as a boundary traction along the right edge of the plate (Neumann loading), while the left edge satisfies homogeneous Dirichlet boundary conditions. The remaining boundaries are traction-free.

Under plane stress assumptions, the constitutive relation between stress and displacement gradients is given by
 
\begin{equation}
    \begin{Bmatrix}
        \sigma_{xx} \\
        \sigma_{yy} \\
        \tau_{xy}
    \end{Bmatrix} = \frac{E}{1 - \nu^{2}}
    \begin{bmatrix}
        1 & \nu & 0 \\
        \nu & 1 & 0 \\
        0 & 0 & \frac{1 - \nu}{2}
    \end{bmatrix} \times
    \begin{Bmatrix}
        \frac{\partial u}{\partial x} \\ 
        \frac{\partial v}{\partial y} \\
        \frac{\partial u}{\partial y} + \frac{\partial v}{\partial x}
    \end{Bmatrix}
\end{equation}
where $E$ is the Young's modulus and $\nu$ is the Poisson ratio of the material. 

The inverse task is to recover the applied boundary force field $\boldsymbol{f}(\boldsymbol{x})$ from observations of the displacement field $(u(\boldsymbol{x}), v(\boldsymbol{x}))$. In this setting, forces are applied along the right edge of the plate and are sampled from a Gaussian random field. We apply traction only in the horizontal direction (i.e., $\boldsymbol{f}=(f_x,0)$ on the loaded boundary), and we condition the sampled force profile to vanish at the endpoints of the loaded boundary, i.e., $ \boldsymbol{f}(1,0)=\boldsymbol{f}(1,1)=\boldsymbol{0}$. Displacements are observed throughout the plate domain.

Because the forward elasticity operator is linear and elliptic, the inverse problem is stable under full-field displacement observations but remains mildly ill-posed due to the smoothing nature of the operator and the boundary-to-interior mapping. As a result, distinct force fields can produce similar displacement patterns, particularly away from the loaded boundary.

\subsection{Chladni Plate Resonance}
\label{sec:chladni}

Chladni patterns arise from resonant vibration: under oscillatory excitation, the displacement field develops structured nodal curves where the response is (approximately) zero~\cite{rossing1982chladni,zhou2016chladni}. A classical model for a thin elastic plate is the Kirchhoff--Love equation~\cite{leissa1969vibration},
\begin{equation}
\rho h\, w_{tt}(x,y,t) + D\,\Delta^{2} w(x,y,t) + \text{(damping)} = f(x,y,t),
\end{equation}
whose eigenmodes and forcing determine the nodal structure. In this benchmark, we approximate plate resonance using a reduced-order modal solver for a simplified second-order PDE on a rectangle. The model retains the key ingredients needed for Chladni-like nodal patterns---mode selectivity, resonance structure, and actuator coupling---while avoiding a full biharmonic plate simulation.

We consider a rectangular domain $\Omega=[0,L]\times[0,M]$ with
$L=M=8.75\times 0.0254$ (meters). We use the cosine eigenfunctions
\begin{equation}
\phi_{nm}(x,y)=\cos(\mu_n x)\cos(\lambda_m y),
\qquad
\mu_n=\frac{n\pi}{L},\ \lambda_m=\frac{m\pi}{M},
\end{equation}
which are compatible with homogeneous Neumann-type boundary conditions (zero normal derivative) on $\partial\Omega$. We truncate to $n=1,\dots,n_{\max}$ and $m=1,\dots,m_{\max}$ with $n_{\max}=m_{\max}=6$. For $n,m\ge 1$, these modes satisfy
$\langle \phi_{nm},\phi_{n'm'}\rangle_{L^2(\Omega)}=\frac{LM}{4}\,\delta_{nn'}\delta_{mm'}$,
which explains the $\frac{4}{LM}$ normalization factors in the modal projections below.

Let $z(x,y,t)$ denote the scalar out-of-plane response. We model the dynamics with a Laplacian-based resonance operator
\begin{equation}
\label{eq:chladni_pde}
z_{tt}(x,y,t) + \big(-\Delta + c_0\big)z(x,y,t) = F(x,y,t),
\qquad (x,y)\in\Omega,
\end{equation}
with homogeneous initial conditions $z(\cdot,0)=0$ and $z_t(\cdot,0)=0$.
The constant shift $c_0$ is chosen as
\begin{equation}
c_0 = 3v^2-\gamma^4,
\qquad v=0.5,\ \gamma=0.02,
\end{equation}
so that the modal frequencies in the cosine basis take the form
\begin{equation}
z_{nm}''(t) + \beta_{nm}^2\,z_{nm}(t)=F_{nm}(t),
\qquad
\beta_{nm}=\sqrt{\mu_n^2+\lambda_m^2+3v^2-\gamma^4}.
\end{equation}
Compared to the Kirchhoff--Love model (with $\Delta^2$), \eqref{eq:chladni_pde} has Laplacian-type dispersion; nevertheless, it preserves a resonance-driven modal decomposition in which nodal patterns arise from interference among a small number of dominant modes.

Each sample is parameterized by random modal amplitudes $\alpha_{nm}\sim 0.01\,\mathcal{N}(0,1)$. The corresponding input field is the modal superposition
\begin{equation}
\label{eq:chladni_input}
S(x,y)=\sum_{n=1}^{n_{\max}}\sum_{m=1}^{m_{\max}}\alpha_{nm}\,\phi_{nm}(x,y),
\end{equation}
which is treated as the \emph{input} function in the dataset. Equivalently, the coefficients can be recovered by orthogonal projection,
\begin{equation}
\alpha_{nm}=\frac{4}{LM}\,\langle S,\phi_{nm}\rangle_{L^2(\Omega)},
\end{equation}
so the forward map can be viewed as operating diagonally in this coefficient representation.

To mimic a localized excitation mechanism, we incorporate a fixed central-actuator coupling at $(x_0,y_0)=(L/2,M/2)$: mode $(n,m)$ is weighted by the mode shape evaluated at the actuator location,
\begin{equation}
\phi_{nm}(x_0,y_0)=\cos\!\Big(\mu_n\frac{L}{2}\Big)\cos\!\Big(\lambda_m\frac{M}{2}\Big).
\end{equation}
We model the time dependence through a prescribed oscillatory envelope with frequency $\omega=500\pi/M$ and evaluate the response at a fixed time $t=t_{\mathrm{fixed}}$ with $t_{\mathrm{fixed}}=6$. The resulting mode-wise transfer coefficient is defined by
\begin{equation}
\label{eq:chladni_timeint}
I_{nm}
=
\int_{0}^{t_{\mathrm{fixed}}}
\sin\!\big(\omega(\tau-t_{\mathrm{fixed}})\big)\,
\exp\!\big(-\gamma^{2}+v^{2}\tau\big)\,
\sin(\beta_{nm}\tau)\,d\tau,
\end{equation}
and we set
\begin{equation}
\label{eq:chladni_transfer}
H_{nm}
=
\frac{4}{LM}\,\phi_{nm}(x_0,y_0)\,\frac{v^2}{\beta_{nm}}\,I_{nm}.
\end{equation}
The output is the fixed-time response field
\begin{equation}
\label{eq:chladni_output}
Z(x,y)=z(x,y,t_{\mathrm{fixed}})
=
\sum_{n=1}^{n_{\max}}\sum_{m=1}^{m_{\max}}
\alpha_{nm}\,H_{nm}\,\phi_{nm}(x,y).
\end{equation}

The inverse map is ill-posed in this setting. Many modal gains $H_{nm}$ can be small due to the central-actuator coupling and the mode-dependent transfer, so recovering $\alpha_{nm}$ from $Z$ amounts to dividing by small numbers, which amplifies noise and numerical error. In addition, $Z$ is a single fixed-time snapshot of a truncated modal expansion, so distinct coefficient sets $\{\alpha_{nm}\}$ can produce nearly indistinguishable response fields, leading to non-uniqueness and multimodality in the inverse direction.

\subsection{Wave Scattering}



This benchmark considers a two-dimensional wave scattering problem in an unbounded domain. The forward problem computes the scattered wavefield given a spatial scatterer configuration using a boundary integral equation method to solve the governing PDEs in unbounded domains. The inverse task is to recover the scatterer geometry from observed wavefields. 

Let $D \subset \mathbb{R}^2$ denote a bounded obstacle with boundary $\partial D$. The scattered field $u^s$ satisfies the Helmholtz equation
\begin{equation}
\label{eq:helmholtz}
\Delta u^s + k^2 u^s = 0,
\end{equation}
in $\mathbb{R}^2 \setminus \overline{D}$,
where $k$ is the wavenumber. The total field $u = u^i + u^s$ satisfies the impedance boundary condition
\begin{equation}
\label{eq:impedance}
\frac{\partial u}{\partial \nu} + i k \lambda u = 0,
\end{equation}
on $\partial D$, where $\nu$ is the unit outward normal vector and $\lambda$ is the impedance function,
and the Sommerfeld radiation condition at infinity. The incident field is taken to be a plane wave $u^i(x) = e^{i k x \cdot d}$.

The inverse task is to infer the obstacle geometry $D$ from observations of the scattered field. Observations consist of far-field patterns sampled at a fixed set of directions. This inverse problem is severely ill-posed due to limited-angle observations, phase ambiguity, and the smoothing nature of wave propagation, leading to non-uniqueness even in the absence of noise.

Scatterer geometries are parameterized in polar coordinates as
\begin{equation}
r(\theta)
=
r_{\mathrm{base}}
\biggl(
1
+
\alpha
\bigl(
0.2 \cos(2\theta)
+
0.15 \sin(3\theta)
+
0.1 \cos(5\theta)
\bigr)
\biggr),
\end{equation}
with $r_{\mathrm{base}} \sim \mathcal{N}(1,0.2)$ and $\alpha \sim \mathcal{N}(0,1)$. Each geometry is rasterized to a $200 \times 200$ binary field that serves as the output representation. The corresponding far-field patterns serve as the inputs.

The forward problem is solved using a boundary integral formulation of the Helmholtz equation with impedance boundary conditions.
For completeness and reproducibility, the full boundary integral derivation and numerical formulation used to generate the dataset are provided below.

We consider the exterior scattering problem. 
The goal is to find the scattered field $u^s \in H^1_{\text{loc}}(\mathbb{R}^2 \setminus \overline{D})$ that satisfies the Helmholtz equation.
The scattered field satisfies the Sommerfeld radiation condition,
\begin{equation}
\label{eq:sommerfeld}
\lim_{r \to \infty}
\sqrt{r}\Big( \frac{\partial u^{s}}{\partial r} - i k u^{s} \Big) = 0.
\end{equation}
Equation $\eqref{eq:helmholtz}$ models time-harmonic propagation in a homogeneous medium. The wavenumber $k$ satisfies $k = 2\pi/\lambda_{\mathrm{wave}}$, where $\lambda_{\mathrm{wave}}$ is the wavelength. Equation \eqref{eq:impedance} represents the impedance boundary condition that models the interaction of waves with the obstacle's surface. Condition \eqref{eq:sommerfeld} ensures that energy radiates outward. We take the incident field to be the plane wave
\begin{equation}
u^{i}(x) = e^{i k, x\cdot d},
\end{equation}
where $d$ is the unit direction vector.

To construct a boundary integral formulation, we utilize the fundamental solution (Green function) of the Helmholtz equation,
\begin{equation}
\Phi(x,y) = \frac{i}{4},H_{0}^{(1)}!\big(k,\lVert x-y\rVert\big),
\end{equation}
where $H_{0}^{(1)}$ is the Hankel function of the first kind of order zero. We represent the scattered field as the single-layer potential
\begin{equation}
u^{s}(x) = \int_{\partial D} \Phi(x,y),\phi(y),ds(y),
\end{equation}
where $x \in \mathbb{R}^{2}\setminus\overline{D}$,
with $\phi$ an unknown boundary density that acts as a continuous distribution of sources along $\partial D$. Once $\phi$ is known, we can calculate the scattered field
anywhere in the domain.

We determine $\phi$ by imposing the impedance boundary condition. Substituting the single-layer representation for $u^{s}$ into \eqref{eq:impedance} yields the boundary integral equation
\begin{equation}
\phi(x) - (K'\phi)(x) - i k \lambda(x),(S \phi)(x)
= 2 \frac{\partial u^{i}}{\partial \nu}(x) - 2i k \lambda(x) u^{i}(x),
\end{equation}
where $x \in \partial D$.
The boundary operators $S$ and $K'$ are defined by
\begin{equation}
(S \phi)(x)
= 2 \int_{\partial D} \Phi(x,y) \phi(y) ds(y),
\end{equation}
and
\begin{equation}
(K'\phi)(x)
= 2 \int_{\partial D}
\frac{\partial \Phi(x,y)}{\partial \nu(x)} \phi(y) ds(y).
\end{equation}

To solve this equation numerically, we parameterize the boundary as
\begin{equation}
\partial D = { z(t) : 0 \le t \le 2\pi }.
\end{equation}
Let $\psi(t)$ denote the resulting transformed boundary density. The far-field pattern takes the form
\begin{equation}
u_{\infty}(\hat{x})
= \frac{e^{-i\pi/4}}{\sqrt{8\pi k}}
\int_{\partial D}
(
k (\nu(y)\cdot \hat{x} )
+ i \lambda(y)
)
e^{-ik \hat{x} \cdot y},
\phi(y) ds(y),
\end{equation}
where $\hat{x} \in S^{1}$ is the observation direction.
In discretized form,
\begin{equation}
u_{\infty}(\hat{x})
= \frac{e^{-i\pi/4}}{\sqrt{8\pi k}}
\sum_{j=0}^{2n-1}
(
k (\nu(z(t_{j}))\cdot \hat{x} )
+ i \mu(t_{j})
)
e^{-ik \hat{x}\cdot z(t_{j})} 
\psi(t_{j}) 
\frac{\pi}{n}.
\end{equation}

\subsection{Full Waveform Inversion}



This benchmark considers a seismic full waveform inversion task adapted from \cite{10.1093/gji/ggac179}. The dataset originates from established simulation pipelines used in seismic inversion studies and is designed to model the recovery of subsurface velocity fields from recorded seismic waveforms.

Each sample consists of a two-dimensional subsurface velocity model defined on a fixed spatial grid and a corresponding set of seismic observations generated by propagating band-limited source wavelets through the medium. Seismic data are recorded at a fixed array of receiver locations for a prescribed acquisition geometry. Both the source and receiver configurations are held constant across all samples.

The dataset is generated by numerically simulating wave propagation through the subsurface models, following the procedures described in \cite{10.1093/gji/ggac179}. The resulting seismic measurements are provided as precomputed waveform data and serve as the outputs of the forward process.
The inverse task is to recover the subsurface velocity field from the recorded seismic observations. 

In our framework, the subsurface velocity fields are treated as input functions, while the seismic observations are represented as fixed-dimensional arrays corresponding to recorded waveforms. We apply our inverse operator models to map from seismic measurements to distributions over plausible velocity fields, allowing us to assess performance in a large-scale, high-dimensional inverse problem representative of practical seismic imaging scenarios.

\section{Supplementary Methods}

\subsection{Inverse Architectures}
\label{section: supplementary methods inverse models}

We extend B2B to inverse neural operator maps operating between coefficient spaces $\beta \in \mathbb{R}^{l}$ and $\alpha \in \mathbb{R}^{k}$. 
The choice of architecture depends on the structure of the inverse problem. Deterministic mappings (linear, nonlinear, and invertible networks) are appropriate when the inverse exhibits unique or nearly unique solutions. Probabilistic mappings (conditional variational autoencoders and mixture density networks) are necessary for ill-posed problems, representing the full distribution $p(\alpha \mid \beta)$ to capture non-uniqueness and uncertainty. We describe deterministic and invertible architectures first, followed by probabilistic methods.

\subsubsection{Linear Inverse Maps} 

We learn linear maps between coefficient vectors with ridge regression using $k=\ell=100$ basis coefficients and $\lambda = 10^{-6}$ regularization.

The linear model learns a matrix $W \in \mathbb{R}^{k \times l}$ that maps output coefficients $\beta$ to input coefficients $\alpha$ via the transformation $\alpha = W\beta$. 
Given training pairs of the form $(\alpha, \beta)$ computed using the pre-trained function encoders, we compute $W$ via $W = A B^\top\,(B B^\top + \lambda I)^{-1}$, where $A=[\alpha^{(1)},\ldots,\alpha^{(N)}]\in\mathbb{R}^{k\times N}$ and $B=[\beta^{(1)},\ldots,\beta^{(N)}]\in\mathbb{R}^{\ell\times N}$. 
At inference time, the model outputs $\hat{\alpha} = W\beta$.

The linear-inverse model first learns a forward linear map $M \in \mathbb{R}^{\ell \times k}$ from input to output coefficients via $\hat{\beta}=M\alpha$ using an analogous ridge objective. At inference time, the model recovers input coefficients by inverting this learned forward map. When $k=\ell$, the model computes $\hat{\alpha}=M^{-1}\beta$ if $M$ is numerically well-conditioned; otherwise, it computes a regularized inverse $\hat{\alpha}=(M^\top M+\lambda I)^{-1}M^\top\beta$.

\subsubsection{Nonlinear Inverse Maps} 

The nonlinear inverse model learns a deterministic mapping from output coefficients to input coefficients using a multilayer perceptron. The model implements a function $f_{\theta} : \mathbb{R}^{\ell} \to \mathbb{R}^{k}$ with three hidden layers of width 512, ReLU activations, and a linear output layer.

Given training pairs $\lbrace (\alpha^{(n)}, \beta^{(n)})\rbrace_{n=1}^{N}$, we train the network by minimizing the mean-squared error objective
$\min_{\theta}\ \sum_{n=1}^{N} \|f_{\theta}(\beta^{(n)}) - \alpha^{(n)}\|_2^2$.
We optimize this objective using gradient descent with the same optimizer and learning rate used for other neural inverse models. At inference time, the model predicts input coefficients via a single forward pass $\hat{\alpha} = f_{\theta}(\beta)$.

\subsubsection{Invertible Neural Networks}
Invertible neural networks (INNs) 
construct bijections using coupling layers, where the input is split into halves that are transformed as functions of each other. We implement both additive coupling transformations and affine variants that incorporate learned scaling for enhanced expressivity. We develop two architectures: standard INNs that directly map coefficient spaces $\alpha \leftrightarrow \beta$, and conditional INNs that map $\alpha \leftrightarrow z$, conditioned on $\beta$ to integrate observational constraints.

\paragraph{Invertible Neural Networks.}
Standard INNs learn a direct bijection between input and output coefficient spaces, $\alpha \leftrightarrow \beta$, through coupling transformations. We use 2 coupling blocks with 3 hidden layers of width 512 each. Each coupling block splits the input into two halves, $x = [x_{1}, x_{2}]$, transforms one half as a function of the other while preserving the other. In the additive variant, the transformation is 
\begin{equation}
    y_{1} = x_{1}, \quad y_{2} = x_{2} + t(x_{1}),
\end{equation}
where $t$ is parameterized by an MLP. In the affine variant, we apply both scaling and translation:
\begin{equation}
    y_{1} = x_{1}, \quad y_{2} = x_{2} \odot \exp(s(x_{1})) + t(x_{1}),
\end{equation}
where $s$ and $t$ are learned functions. The roles of the two halves alternate across successive blocks, ensuring all dimensions undergo nonlinear transformations. The inverse is computed analytically by reversing these operations without numerical approximation. For affine couplings, the log-determinant of the Jacobian is
\begin{equation}
    \log|\det J| = \sum_{i} s_{i},
\end{equation}
which enables tractable density evaluation. Training proceeds bidirectionally: we minimize reconstruction error in both forward ($\alpha \to \beta$) and inverse ($\beta \to \alpha$) directions using shared parameters, providing mutual regularization. The bidirectional loss is
\begin{equation}
    \mathcal{L} = \lVert \beta - f_{\theta}(\alpha) \rVert^{2} + \lVert \alpha - f_{\theta}^{-1}(\beta) \rVert^{2}.
\end{equation}

\paragraph{Conditional Invertible Neural Networks (Deterministic).}

We implement a deterministic conditional invertible neural network (cINN) to map output coefficients $\beta$ to input coefficients $\alpha$ using coupling-based invertible transformations conditioned on $\beta$. The model introduces an auxiliary latent vector $z$ only to define an invertible transformation; the model does not treat $z$ as a random variable and does not impose a latent prior.

Each coupling block conditions its transformation on $\beta$ by providing $\beta$ as input to the coupling networks. In the additive coupling variant, a block applies
\begin{equation}
    y_1 = x_1,\qquad y_2 = x_2 + t(x_1,\beta),
\end{equation}
and in the affine coupling variant, a block applies
\begin{equation}
    y_1 = x_1,\qquad y_2 = x_2 \odot \exp(s(x_1,\beta)) + t(x_1,\beta),
\end{equation}
where $s(\cdot,\beta)$ and $t(\cdot,\beta)$ are multilayer perceptrons.

We train the deterministic cINN on the inverse mapping by fixing $z=0$ and minimizing mean-squared error in coefficient space. Given coefficient pairs $\lbrace(\alpha^{(n)},\beta^{(n)})\rbrace_{n=1}^{N}$, we compute predictions $\hat{\alpha}=f_\theta^{-1}(z=0,\beta)$ and minimize
$\mathcal{L}_{\mathrm{cINN}} = \|\hat{\alpha}-\alpha\|_2^2$.
We apply no latent regularizer to avoid introducing conflicting objectives in this deterministic setting. At inference time, the model outputs $\hat{\alpha}=f_\theta^{-1}(0,\beta)$ using the same fixed latent vector.

\paragraph{Conditional Invertible Neural Networks (Probabilistic).}
We also consider a probabilistic variant of the conditional invertible neural network that explicitly models the conditional density $p(\alpha \mid \beta)$ in coefficient space. This model uses the same coupling-based invertible architecture as the deterministic cINN but introduces stochasticity through a latent variable and trains by maximum likelihood.

The probabilistic cINN defines an invertible transformation between input coefficients $\alpha$ and a latent variable $z$, conditioned on output coefficients $\beta$,
\begin{equation}
z = f_\theta(\alpha; \beta), \qquad \alpha = f_\theta^{-1}(z; \beta),
\end{equation}
where $z \sim \mathcal{N}(0,I)$. Each coupling block conditions its transformation on $\beta$ by providing $\beta$ as input to the coupling networks. We use additive coupling layers, optionally interleaved with ActNorm layers that introduce non-zero log-determinant contributions.

We train the probabilistic cINN by maximizing the conditional log-likelihood of the observed coefficients under the change-of-variables formula. Given training pairs $\lbrace (\alpha^{(n)},\beta^{(n)})\rbrace_{n=1}^{N}$, the negative log-likelihood objective takes the form
$\frac{1}{2}\,\mathbb{E}\big[\|z\|_2^2\big]
- \mathbb{E}\big[\log |\det J|\big]$,
where $z = f_\theta(\alpha;\beta)$ and $\log |\det J|$ denotes the log-determinant of the Jacobian of the transformation. For additive coupling layers, the Jacobian determinant is zero; ActNorm layers contribute the non-trivial log-determinant terms required for density estimation.

At inference time, given output coefficients $\beta$, the model generates diverse inverse predictions by sampling $z \sim \mathcal{N}(0,I)$ and computing $\hat{\alpha} = f_\theta^{-1}(z;\beta)$. This procedure yields samples from the learned conditional distribution $p(\alpha \mid \beta)$.

\subsubsection{Conditional Variational Autoencoder}

The cVAE is composed of two networks: an encoder and a decoder.
The encoder learns a distribution over a latent embedding space $\mathcal{Z}$ that is conditioned on both the input coefficients $\alpha$ and the corresponding output space coefficients $\beta$,
\begin{equation}
    p_{\theta}(z \mid \alpha, \beta) = \mathcal{N}(\mu_{\theta}(\alpha, \beta), \Sigma_{\theta}(\alpha, \beta)),
\end{equation}
where $\mu_{\theta}$ and $\Sigma_{\theta}$ are parameterized by neural networks.
Intuitively, the latent space distribution captures the statistical properties of the joint distribution over coefficients $\alpha$ and $\beta$.
The decoder is a generative model. 
For a given $\beta \in \mathbb{R}^{l}$, the reconstructed coefficients $\hat{\alpha}$ are sampled from the distribution,
\begin{equation}
    \hat{\alpha} \sim p_{\theta}(\alpha \mid \beta, z),
\end{equation}
where $z \sim p(z)$, which is typically $\mathcal{N}(0, I)$.

We train the cVAE by maximizing the evidence lower bound (ELBO), 
\begin{equation}
    - \mathbb{E}_{z \sim p_{\theta}(z \mid \alpha, \beta)} [\log p_{\theta}(\alpha \mid \beta, z)] + \mathrm{KL}(p_{\theta}(z \mid \alpha, \beta) \,\|\, p(z)).
\end{equation}
The first loss term is the negative log likelihood, which acts as a reconstruction loss in the input space. The second term is the Kullback-Leibler (KL) divergence that regularizes the learned posterior distribution toward the prior distribution. 
To further stabilize training, we add a forward consistency loss that enforces compatibility between reconstructed coefficients and the forward operator, $\mathbb{E}_{z \sim p_{\theta}(z \mid \alpha, \beta)} [\lVert \beta - T(\hat{\alpha}) \rVert^{2}]$, where $\hat{\alpha}$ is the reconstructed input coefficients.


\subsubsection{Conditional Normalizing Flow (RealNVP)}
We implement a conditional normalizing flow based on the RealNVP architecture to model the conditional density $p(\alpha \mid \beta)$ in coefficient space. The model defines an invertible transformation between input coefficients $\alpha \in \mathbb{R}^{k}$ and a latent variable $z \in \mathbb{R}^{k}$, conditioned on output coefficients $\beta$,
\begin{equation}
z = f_\theta(\alpha;\beta), \qquad \alpha = f_\theta^{-1}(z;\beta),
\end{equation}
with a standard Gaussian base distribution $z \sim \mathcal{N}(0,I)$.

The transformation $f_\theta$ consists of a sequence of conditional affine coupling layers with alternating masks. Each coupling layer conditions its scale and translation functions on both the masked coefficients and the conditioning variable $\beta$. We bound the scaling functions to ensure numerical stability. All transformations admit closed-form inverses, and the Jacobian determinant is tractable by construction.

We train the model by maximizing the exact conditional log-likelihood under the change-of-variables formula,
\begin{equation}
\log p_\theta(\alpha \mid \beta)
=
\log p_0(f_\theta(\alpha;\beta))
+
\log \left|\det \frac{\partial f_\theta(\alpha;\beta)}{\partial \alpha}\right|,
\end{equation}
using paired coefficient data $(\alpha,\beta)$.

At inference time, given output coefficients $\beta$, we sample $z \sim \mathcal{N}(0,I)$ and generate inverse predictions via $\hat{\alpha} = f_\theta^{-1}(z;\beta)$. We decode the resulting coefficients into input functions using the learned input basis.

\subsubsection{Mixture Density Networks}
Mixture density networks (MDNs) directly predict a mixture of Gaussians over input coefficients conditioned on output observations. The architecture consists of an MLP that maps $\beta \to \lbrace \pi_{k}, \mu_{k}, \Sigma_{k} \rbrace_{k=1}^{K}$, where $\pi_{k}$ are mixture weights, $\mu_{k} \in \mathbb{R}^{k}$ are component means, and $\Sigma_{k}$ are precision matrices. For computational efficiency, we parameterize precisions as diagonal matrices, expressed via learned log-precisions that are exponentiated to ensure positivity. The predicted distribution is 
\begin{equation}
    p(\alpha \mid \beta) = \sum_{k=1}^{K} \pi_{k} \mathcal{N}(\alpha \mid \mu_{k}, \Sigma_{k}^{-1}).
\end{equation}
Training minimizes the negative log-likelihood of observed input coefficients under the predicted mixture. For each component, the log-probability is computed as
\begin{equation}
    \log p_{k}(\alpha) = -\frac{1}{2}(\alpha - \mu_{k})^{\top}\Sigma_{k}(\alpha - \mu_{k}) - \frac{d}{2}\log(2\pi) + \frac{1}{2}\log|\Sigma_{k}|,
\end{equation}
where $d$ is the dimensionality of $\alpha$. The mixture log-likelihood is then
\begin{equation}
    \log p(\alpha \mid \beta) = \log\sum_{k=1}^{K} \pi_{k} \exp(\log p_{k}(\alpha)),
\end{equation}
computed via the log-sum-exp operation for numerical stability. At inference, we sample a mixture component according to $\pi_{k}$ and draw from the corresponding Gaussian. 

\subsection{Invertible Fourier Neural Operators}

For comparison, we implement an invertible Fourier neural operator (iFNO). The iFNO is a unified architecture that co-learns forward and inverse mappings with shared parameters, so information flows in both directions and acts as mutual regularization for ill-posed, noise-sensitive inverse inference while preserving the global modeling capacity of Fourier layers. iFNO stacks invertible Fourier blocks in a lifted latent channel space: a pointwise projector $P$ lifts the discretized input and splits the channels into two halves; within each block, one half is transformed by a Fourier layer $L$, passed through a strictly positive softplus-based scaling $S$, and used via element-wise multiplication to update the other half, after which the roles swap. Because the updates are multiplicative with positive scales and each sub-operation is invertible, every block is a bijection and the stack is exactly reversible by processing blocks in reverse and applying reciprocal scalings. For forward prediction, the final halves are concatenated and mapped to the output by $Q$. For inverse prediction, $P'$ lifts observed outputs into the latent space at the last block, the blocks are inverted sequentially to recover the first-block inputs, and $Q'$ maps them back to the input discretization.

To enhance identifiability and enable posterior inference, iFNO integrates a $\beta$-VAE on the recovered inputs, using an encoder-decoder to reconstruct input functions and produce uncertainty estimates useful for inverse tasks. Training proceeds in three phases: supervised learning of the invertible stack with forward and inverse objectives augmented by projection-consistency terms that stabilize $P/Q$ and $P'/Q'$; standalone optimization of the $\beta$-VAE with a variational objective; and joint end-to-end fine-tuning of all components, which empirically improves stability and accuracy in both directions \cite{pmlr-v258-long25a}.

\end{document}